\documentclass{article}

\usepackage[preprint,nonatbib]{corl_2022} 

\usepackage{microtype}
\usepackage{graphicx}
\usepackage{booktabs} 

\usepackage{hyperref}

\usepackage{amsmath}
\usepackage{amssymb}
\usepackage{mathtools}
\usepackage{amsthm}
\usepackage{scalerel}
\usepackage{makecell}
\usepackage{bm}
\usepackage{multirow}
\usepackage{wrapfig}

\usepackage[natbib=true,style=numeric,sorting=none,backend=biber,maxcitenames=2,maxbibnames=99,giveninits]{biblatex}
\usepackage{siunitx}
\usepackage{algorithm}
\usepackage{algorithmic}

\usepackage{pgf, tikz}
\usetikzlibrary{arrows, automata}

\usepackage{subcaption}

\usepackage[capitalize,noabbrev]{cleveref}

\theoremstyle{plain}

\theoremstyle{definition}

\theoremstyle{remark}

\newcommand\given[1][]{\:#1\vert\:}
\newcommand{\mi}{\mathcal{I}}

\newcommand{\exn}{\mathbb{E}}

\newcommand{\agent}{\texttt{SEADS}\xspace}

\usepackage{xspace}
\newcommand{\LightsOut}{\emph{LightsOut}\xspace}
\newcommand{\TileSwap}{\emph{TileSwap}\xspace}
\newcommand{\Cursor}{\emph{Cursor}\xspace}
\newcommand{\Reacher}{\emph{Reacher}\xspace}
\newcommand{\Jaco}{\emph{Jaco}\xspace}
\newcommand{\LightsOutReacher}{\texttt{LightsOutReacher}\xspace}
\newcommand{\TileSwapReacher}{\texttt{TileSwapReacher}\xspace}
\newcommand{\LightsOutJaco}{\texttt{LightsOutJaco}\xspace}
\newcommand{\TileSwapJaco}{\texttt{TileSwapJaco}\xspace}
\newcommand{\LightsOutCursor}{\texttt{LightsOutCursor}\xspace}
\newcommand{\TileSwapCursor}{\texttt{TileSwapCursor}\xspace}

\usepackage[textsize=tiny]{todonotes}

\title{Learning Temporally Extended Skills in Continuous Domains as Symbolic Actions for Planning}

\author{
  Jan Achterhold \hspace{7ex} Markus Krimmel \hspace{7ex}  Joerg Stueckler\\
  Embodied Vision Group, Max Planck Institute for Intelligent Systems,
  T\"ubingen, Germany \\
  \texttt{\{jan.achterhold,markus.krimmel,joerg.stueckler\}@tuebingen.mpg.de} \\
}

\begin{document}
\maketitle


\begin{abstract}
Problems which require both long-horizon planning and continuous control capabilities pose significant challenges to existing reinforcement learning agents. In this paper we introduce a novel hierarchical reinforcement learning agent which links temporally extended skills for continuous control with a forward model in a symbolic discrete abstraction of the environment's state for planning. We term our agent \agent for Symbolic Effect-Aware Diverse Skills.
We formulate an objective and corresponding algorithm which leads to
unsupervised learning of a diverse set of skills through intrinsic motivation given a known state abstraction.
The skills are jointly learned with the symbolic forward model which captures the effect of skill execution in the state abstraction.
After training, we can leverage the skills as symbolic actions using the forward model for long-horizon planning
and subsequently execute the plan using the learned continuous-action control skills.
The proposed algorithm learns skills and forward models that can be used to solve complex tasks which
require both continuous control and long-horizon planning capabilities with high success rate.
It compares favorably with other flat and hierarchical reinforcement learning baseline agents and is successfully demonstrated with a real robot. Project page: \url{https://seads.is.tue.mpg.de}
\end{abstract}



\section{Introduction}

Reinforcement learning (RL) agents have been applied to difficult continuous control and discrete planning problems such as the DeepMind Control Suite~\cite{tassa2018dmcontrolsuite}, StarCraft II~\cite{vinyals2019_alphastar}, or Go~\cite{silver2016_alphago} in recent years.
Despite this tremendous success, tasks which require both continuous control capabilities and long-horizon discrete planning are classically approached with task and motion planning~\cite{garrett2021_tamp}. 
These problems still pose significant challenges to RL agents~\cite{mirza2020_physembplan}.
An exemplary class of environments which require both continuous-action control and long-horizon planning are \emph{physically embedded games} as introduced by \cite{mirza2020_physembplan}. In these environments, a board game is embedded into a physical manipulation setting. A move in the board game can only be executed indirectly through controlling a physical manipulator such as a robotic arm. We simplify the setting of \cite{mirza2020_physembplan} and introduce physically embedded \emph{single-player} board games which do not require to model the effect of an opponent.
Our experiments support the findings of \cite{mirza2020_physembplan} that these environments are challenging to solve for existing flat and hierarchical RL agents.
In this paper, we propose a novel hierarchical RL agent for such environments which learns skills and their effects in a known symbolic abstraction of the environment.

\begin{figure}[tb!]
    \captionsetup[subfigure]{labelformat=empty}
    \begin{subfigure}[t]{0.13\linewidth}
    \centering
    \includegraphics[width=\linewidth]{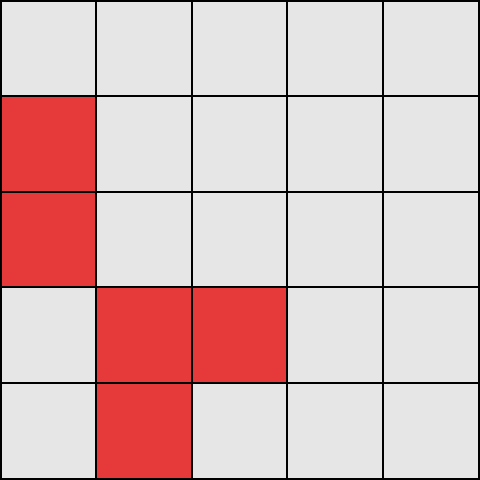}
    \caption{\scriptsize\LightsOutCursor}
    \label{fig:lightsout_board}
    \end{subfigure}\hfill
    \begin{subfigure}[t]{0.13\linewidth}
    \centering
    \includegraphics[width=\linewidth]{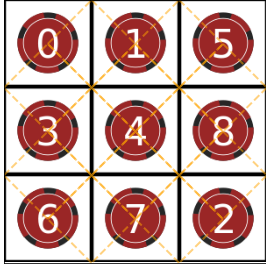}
    \caption{\scriptsize\TileSwapCursor}
    \label{fig:tileswap_board}
    \end{subfigure}\hfill
    \begin{subfigure}[t]{0.13\linewidth}
    \centering
    \includegraphics[width=\linewidth]{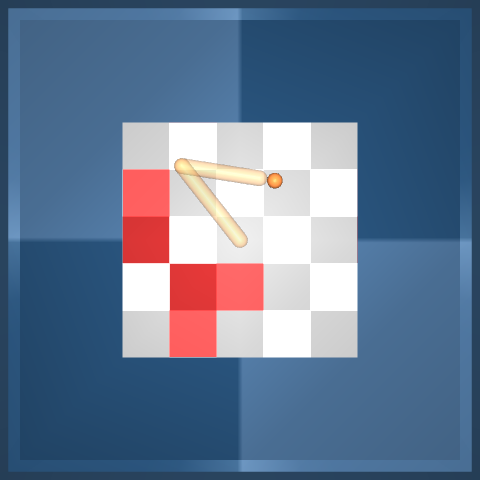}
    \caption{\scriptsize\LightsOutReacher}
    \label{fig:lightsoutreacher}
    \end{subfigure}\hfill
    \begin{subfigure}[t]{0.13\linewidth}
    \centering
    \includegraphics[width=\linewidth]{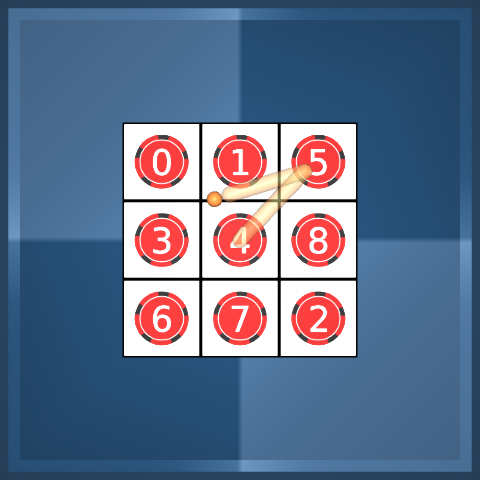}
    \caption{\scriptsize\TileSwapReacher}
    \end{subfigure}\hfill
    \begin{subfigure}[t]{0.13\linewidth}
    \includegraphics[width=\linewidth]{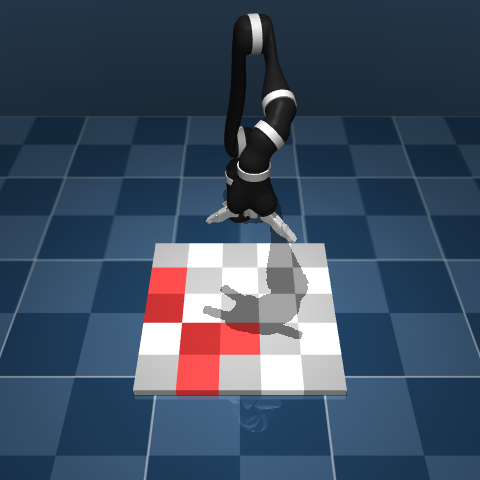}
    \caption{\scriptsize\LightsOutJaco}
    \label{fig:lightsoutjaco}
    \end{subfigure}\hfill
    \begin{subfigure}[t]{0.13\linewidth}
    \includegraphics[width=\linewidth]{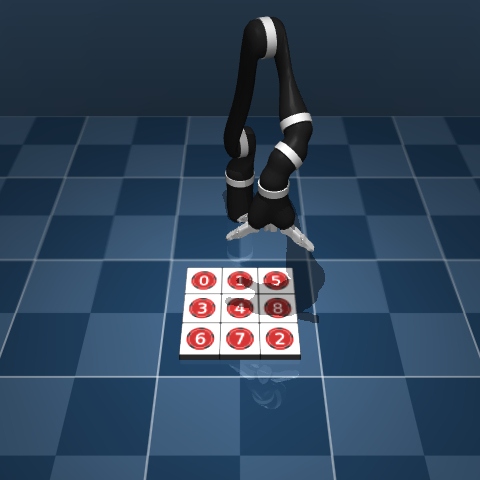}
    \caption{\scriptsize\TileSwapJaco}
    \end{subfigure}
    \caption{\LightsOut (with on (red) and off (gray) fields) and \TileSwap (fields in a rhombus are swapped if pushed inside) board games embedded into physical manipulation settings. A move in the board game can only indirectly be executed through controlling a manipulator.}
    \label{fig:embedded_environments}
\end{figure}

As a concrete example for a proposed embedded single-player board game we refer to the \emph{LightsOutJaco} environment (see Fig.~\ref{fig:embedded_environments}). Pushing a field on the \emph{LightsOut} board toggles the illumination state (\emph{on} or \emph{off}) of the field and its non-diagonal neighboring fields. A field on the board can only be pushed by the end effector of the \emph{Jaco} robotic arm. The goal is to reach a board state in which all fields are \emph{off}.
The above example also showcases the two concepts of \emph{state} and \emph{action} abstraction in decision making~\cite{konidaris2019_abstraction}. A state abstraction function $\Phi(s_t)$ only retains information in state $s_t$ which is relevant for a particular decision making task. In the LightsOut example, to decide which move to perform next (i.e., which field to push), only the illumination state of the board is relevant. A \emph{move} can be considered an \emph{action abstraction}: A skill, i.e. high-level action (e.g., push top-left field), comprises a sequence of low-level actions required to control the robotic manipulator. 

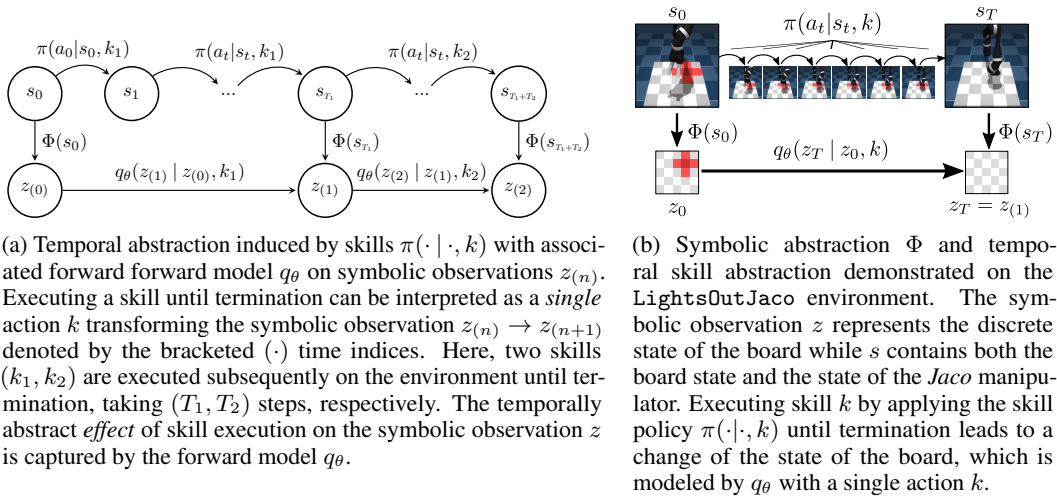
\begin{figure*}[t!]
\centering
\begin{subfigure}[t]{0.57\linewidth}
\centering
\resizebox{\textwidth}{!}{
\begin{tikzpicture}[
        > = stealth, 
        shorten > = 1pt, 
        auto,
        node distance = 1.8cm, 
        semithick 
    ]

    \tikzstyle{every state}=[
        draw = black,
        thick,
        fill = white,
        minimum size = 10mm,
        scale = 1
    ]
    
    \node[state] (s0) {$s_0$};
    \node[state] (s1) [right of = s0] {$s_1$};
    \node (sd1) [right of = s1] {$...$};
    \node[state] (st1) [right of = sd1] {$s_{\scaleto{T_1}{3.5pt}}$};
    \node (sd2) [right of = st1] {$...$};
    \node[state] (st12) [right of = sd2] {$s_{\scaleto{T_1+T_2}{3.5pt}}$};
    
    \path[->] (s0) edge [bend left=40] node {$\pi(a_0|s_0,k_1)$} (s1);
    \path[->] (s1) edge [bend left=40] node {$\pi(a_t|s_t,k_1)$} (sd1);
    \path[->] (sd1) edge [bend left=40] (st1);
    \path[->] (st1) edge [bend left=40] node {$\pi(a_t|s_t,k_2)$} (sd2);
    \path[->] (sd2) edge [bend left=40] (st12);
    
    \node[state] (z0) [below of = s0] 
    {$z_{(0)}$};
    \node[state] (z1) [below of = st1] 
    {$z_{(1)}$};
    \node[state] (z2) [below of = st12] 
    {$z_{(2)}$};
    
    \path[->] (z0) edge node {$q_\theta(z_{(1)} \given z_{(0)},k_1)$} (z1);
    \path[->] (z1) edge node {$q_\theta(z_{(2)} \given z_{(1)},k_2)$} (z2);
    
    \path[->] (s0) edge node {$\Phi(s_0)$} (z0);
    \path[->] (st1) edge node {$\Phi(s_{\scaleto{T_1}{3.5pt}})$} (z1);
    \path[->] (st12) edge node {$\Phi(s_{\scaleto{T_1+T_2}{3.5pt}})$} (z2);
    
    \end{tikzpicture}
    }
    \caption{Temporal abstraction induced by  skills $\pi(\cdot \given \cdot, k)$ with associated forward forward model $q_\theta$ on symbolic observations $z_{(n)}$. Executing a skill until termination can be interpreted as a \emph{single} action $k$ transforming the symbolic observation $z_{(n)} \rightarrow z_{(n+1)}$ denoted by the bracketed $(\cdot)$ time indices. Here, two skills $(k_1, k_2)$ are executed subsequently on the environment until termination, taking $(T_1, T_2)$ steps, respectively. The temporally abstract \emph{effect} of skill execution  on the symbolic observation $z$ is captured by the forward model $q_\theta$. }
    \label{fig:overview_a}

\end{subfigure} \hfill
\begin{subfigure}[t]{0.4\linewidth}
\centering
\includegraphics[width=0.99\linewidth]{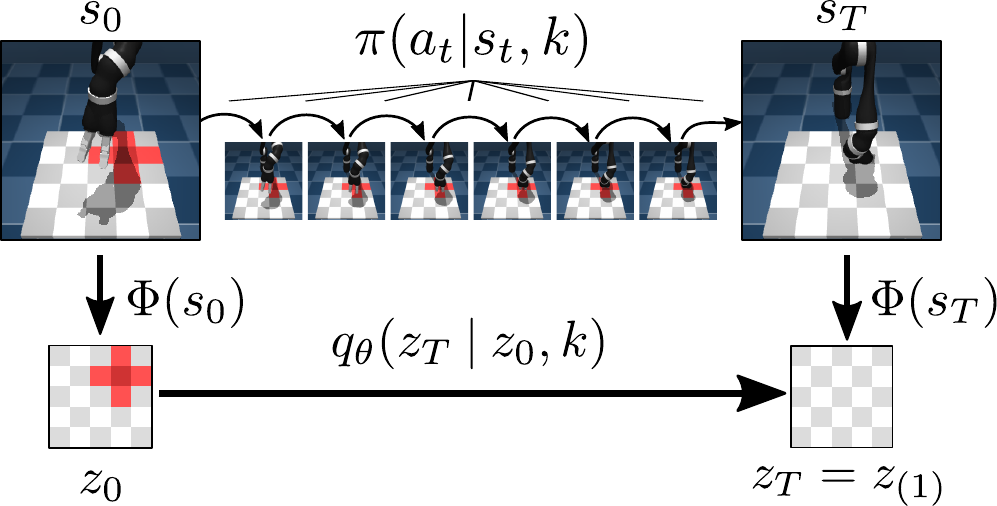}
\caption{Symbolic abstraction $\Phi$ and temporal skill abstraction demonstrated on the \LightsOutJaco environment. The symbolic observation $z$ represents the discrete state of the board while $s$ contains both the board state and the state of the \Jaco manipulator. Executing skill $k$ by applying the skill policy $\pi(\cdot | \cdot, k)$ until termination leads to a change of the state of the board, which is modeled by $q_\theta$ with a single action $k$.}
\end{subfigure}
\caption{We aim to learn skills with associated policies $\pi(a_t | s_t, k)$ which lead to diverse and predictable (by the forward model $q_\theta$) transitions in a symbolic abstraction $z = \Phi(s)$ of the state $s$.}
\label{fig:overview}
\end{figure*}

We introduce a two-layer hierarchical agent which assumes a discrete state abstraction $z_t = \Phi(s_t) \in \mathcal{Z}$ to be known and observable in the environment, which we in the following refer to as \emph{symbolic observation}. 
In our approach, we assume that state abstractions can be defined manually for the environment. 
For LightsOut, the symbolic observation corresponds to the state of each field (on/off). 
We provide the state abstraction as prior knowledge about the environment and assume that skills induce changes of the abstract state.
Our approach then learns a diverse set of skills for the given state abstraction as action abstractions and a corresponding forward model which predicts the effects of skills on abstract states.
In board games, these abstract actions relate to \emph{moves}. 
We jointly learn the predictive forward model $q_\theta$ and \emph{skill policies} $\pi(a \given s_t, k)$ for low-level control through an objective which maximizes the number of symbolic states reachable from any state of the environment (diversity) and the predictability of the effect of skill execution.
Please see Fig.~\ref{fig:overview} for an illustration of the introduced temporal and symbolic hierarchy.
 The forward model $q_\theta$ can be leveraged to plan a sequence of skills to reach a particular state of the board (e.g., all fields off), i.e. to solve tasks. 
We evaluate our approach using two single-player board games in environments with varying complexity in continuous control. We demonstrate that our agent learns skill policies and forward models suitable for solving the associated tasks with high success rate and compares favorably with other flat and hierarchical RL baseline agents. We also demonstrate our agent playing LightsOut with a real robot.

In summary, we contribute the following: (1) We formulate a novel RL algorithm which, based on a state abstraction of the environment and an information-theoretic objective, jointly learns a diverse set of continuous-action skills and a forward model capturing the temporally abstracted effect of skill execution in symbolic states.
    (2) We demonstrate the superiority of our approach compared to other flat and hierarchical baseline agents in solving complex physically-embedded single-player games, requiring high-level planning and continuous control capabilities.
We provide additional materials, including video and code, at \url{https://seads.is.tue.mpg.de}.

\section{Related work}

{\bf Diverse skill learning and skill discovery.}
Discovering general skills to control the environment through exploration without task-specific supervision is a fundamental challenge in RL research.
DIAYN~\cite{eysenbach2019diversity} formulates skill discovery using an information-theoretic objective as reward.
The agent learns a skill-conditioned policy for which it receives reward if the target states can be well predicted from the skill.
VALOR~\cite{achiam2018_valor} proposes to condition the skill prediction model on the complete trajectory of visited states.
\citet{warde-farley2019_discern} train a goal-conditioned policy to reach diverse states in the environment.
Variational Intrinsic Control~\cite{gregor2017_vic} proposes to use an information-theoretic objective to learn a set of skills which can be identified from their initial and target states.
Relative Variational Intrinsic Control~\cite{baumli2021_rvic} seeks to learn skills relative to their start state, aiming to avoid skill representations that merely tile the state space into goal state regions.
Both approaches do not learn a forward model on the effect of skill execution like our approach.
\citet{sharma2020dynamics} propose a model-based RL approach (DADS) which learns a set of diverse skills and their dynamics models using mutual-information-based exploration. 
While DADS learns skill dynamics as immediate behavior $q(s_{t+1} | s_t, k)$, we learn a transition model on the effect of skills $q(z_T | z_0, k)$ in a symbolic abstraction, thereby featuring temporal abstraction.

{\bf Hierarchical RL.}
Hierarchical RL can overcome sparse reward settings and time extended tasks by breaking the task down into subtasks.
Some approaches such as methods based on MAXQ~\cite{dietterich2000_maxq,li2017_csrl} assume prior knowledge on the task-subtask decomposition.
In SAC-X~\cite{riedmiller2018_sacx}, auxiliary tasks assist the agent in learning sparse reward tasks and hierarchical learning involves choosing between tasks.
\citet{florensa2017stochastic} propose to learn a span of skills using stochastic neural networks for representing policies. 
The policies are trained in a task-agnostic way using a measure of skill diversity based on mutual information.
Specific tasks are then tackled by training an RL agent based on the discovered skills.
Feudal approaches~\cite{dayan1992feudal} such as HIRO~\cite{nachum2018data} and HAC~\cite{levy2019_hac} train a high-level policy to provide subgoals for a low-level policy.
In our method, we impose that a discrete state-action representation exists in which learned skills are discrete actions, and train the discrete forward model and the continuous skill policies jointly.
Several approaches to hierarchical RL are based on the options framework~\cite{sutton1999mdps} which learns policies for temporally extended actions in a two-layer hierarchy.
Learning in the options framework is usually driven by task rewards. 
Recent works extend the framework to continuous spaces and discovery of options (e.g.~\cite{bacon2016option,bagaria2020_dsc}).
HiPPO~\cite{li2020_hippo} develop an approximate policy gradient method for hierarchies of actions.
HIDIO~\cite{zhang2021_hidio} learns task-agnostic options using a measure of diversity of the skills.
In our approach, we also learn task-agnostic (for the given state abstraction) hierarchical representations using a measure of intrinsic motivation.
However, an important difference is that we do not learn high-level policies over options using task rewards, but learn a skill-conditional forward model suitable for planning to reach a symbolic goal state. 
Jointly, continuous policies are learned which implement the skills.
{Several approaches combine symbolic planning in a given domain description (state and action abstractions) with RL to execute the symbolic actions~\cite{ryan2002_planrl,lyu2019_sdrl,illanes2020_symbplans,kokel2021_reprel,guan2022_skilldiv}.
Similar to our approach, the work in \citet{guan2022_skilldiv} learns low-level skill policies using an information-theoretic diversity measure which implement known symbolic actions.
Differently, we learn the action abstraction and low-level skills given the state abstraction.}

{\bf Representation learning for symbolic planning.}
Some research has been devoted to learning representations for symbolic planning.
\citet{konidaris2018_skillstosymbols} propose a method for acquiring a symbolic planning domain from a set of low-level options which implement abstract symbolic actions.
In~\cite{james2020_portable} the approach is extended to learning symbolic representations for families of SMDPs which describe options in a variety of tasks.
Our approach learns action abstractions as a set of diverse skills given a known state abstraction and a termination condition which requires abstract actions to change abstract states.
\citet{icarte2019_rewardmachines} learn structure and transition models of finite state machines through RL.
\citet{ugur2015_effects} acquire symbolic forward models for a predefined low-level action repertoire in a robotic manipulation context.
{\citet{chitnis2022_nsrt} concurrently learn transition models on the symbolic and low levels from demonstrations provided in the form of hand-designed policies, and use the learned models for bilevel task and motion planning. 
The approach also assumes the state abstraction function to be known.
In \cite{silver2022_nssbilevel} a different setting is considered in which the symbolic transition model is additionally assumed known and skill policies that execute symbolic actions are learned from demonstrations.}
Other approaches such as DeepSym~\cite{ahmetoglu2020_deepsim} or LatPlan~\cite{asai2018classical} learn mappings of images to symbolic states and learn action-conditional forward models. In~\cite{asai2018classical} symbolic state-action representations are learned from image observations of discrete abstract actions (e.g. moving puzzle tiles to discrete locations) which already encode the planning problem. 
Our approach concurrently learns a diverse set of skills (discrete actions) based on an information-theoretic intrinsic reward and the symbolic forward model. 
Differently, in our approach low-level actions are continuous.

\section{Method}
\label{sec:method}

Our goal is to learn a hierarchical RL agent which (i) enables high-level, temporally abstract planning to reach a particular goal configuration of the environment (as given by a symbolic observation) and (ii) features continuous control policies to execute the high-level plan.
Let $\mathcal{S}, \mathcal{A}$ denote the state and action space of an environment, respectively. 
In general, by $\mathcal{Z} = \{0, 1\}^D$ we denote the space of discrete symbolic environment observations $z \in \mathcal{Z}$ and assume the existence of a state abstraction $\Phi: \mathcal{S} \rightarrow \mathcal{Z}$. 
The dimensionality of the symbolic observation $D$ is environment-dependent. 
For the \emph{LightsOutJaco} environment, the state $s = [q, \dot{q}, z] \in \mathcal{S}$ contains the robot arms' joint positions and velocities $(q, \dot{q})$ and a binary representation of the board $z \in \{0, 1\}^{5\times5}$. 
The action space $\mathcal{A}$ is equivalent to the action space of the robotic manipulator. 
In the \emph{LightsOutJaco} example, it contains the target velocity of all actuable joints. { The discrete variable $k \in \mathcal{K} = \{1, ..., K\}$ refers to a particular \emph{skill}, which we will detail in the following. The number of skills $K$ needs to be set in advance, but can be chosen larger than the number of actual skills.}

We equip our agent with symbolic planning and plan execution capabilities through two components: 
{ First, a forward model $\hat{z} = f(z, k) = \operatorname{argmax}_{z'} q_\theta(z' \given z, k)$ allows to \emph{enumerate} all possible symbolic successor states $\hat{z}$ of the current symbolic state $z$ by iterating over the discrete variable $k$.
}
This allows for node expansion in symbolic planners. 
Second, a family of discretely indexed policies $\pi: \mathcal{A}\times\mathcal{S}\times\mathcal{K} \rightarrow \mathbb{R}, \: a_t \sim \pi(a_t \given s_t, k)$ aims to steer the environment into a target state $s_T$ for which it holds that $\Phi(s_T) = \hat{z}$, given that $\Phi(s_0) = z$ and $\hat{z} = \operatorname{argmax}_{z'} q_\theta(z' \given z, k)$ (see Fig.~\ref{fig:overview}). We can relate this discretely indexed family of policies to a set of $K$ \emph{options} \cite{sutton1999mdps}. An option is formally defined as a triple $o_k = (I_k, \beta_k, \pi_k)$ where $I_k \subseteq \mathcal{S}$ is the set of states in which option $k$ is applicable, $\beta_k: \mathcal{S}\times\mathcal{S} \rightarrow [0, 1]$, $\beta_k(s_0,s_t)$ parametrizes a Bernoulli probability of termination in state $s_t$ when starting in $s_0$ (f.e. when detecting an abstract state change) and $\pi_k(a \given s_t): \mathcal{S} \rightarrow \Delta(\mathcal{A})$ is the option policy on the action space $\mathcal{A}$. We will refer to the option policy as \emph{skill policy} in the following.
We assume that all options are applicable in all states, i.e., $I_k = \mathcal{S}$. An option terminates if the symbolic state has changed between $s_0$ and $s_t$ or a timeout is reached, i.e., $\beta_k(s_0,s_t) = \mathbf{1}[\Phi(s_0) \neq \Phi(s_t) \vee t = t_{\max} ]$.
To this end, we append a normalized counter $t/t_{\max}$ to the state $s_t$.
We define the operator $\operatorname{apply}$ as 
$s_T = \operatorname{apply}(E, \pi, s_0, k)$    
which applies the skill policy $\pi(a_t \given s_t, k)$ until termination on environment $E$ starting from initial state $s_0$ and returns the terminal state $s_T$.
We also introduce a bracketed time notation which abstracts the effect of skill execution from the number of steps $T$ taken until termination 
    $s_{(n)} = \operatorname{apply}(E, \pi, s_{(n-1)}, k)$
with $n \in \mathbb{N}_0$ (see Fig.~\ref{fig:overview_a}). The $\operatorname{apply}$ operator can thus be rewritten as $s_{(1)} = \operatorname{apply}(E, \pi, s_{(0)}, k)$ with $s_{(0)} = s_0$, $s_{(1)} = s_T$.
%
The symbolic forward model $q_\theta(z_T \given z_0, k)$
aims to capture the relation of $z_0$, $k$ and $z_T$ for $s_T = \operatorname{apply}(E, \pi, s_0, k)$ with $z_0=\Phi(s_0), z_T=\Phi(s_T)$. 
The model factorizes over the symbolic observation as $q_\theta(z_T \given k, z_0) = \prod_{d=1}^D q_\theta([z_T]_{d} \given k, z_0) = \prod_{d=1}^D \mathrm{Bernoulli}([\alpha_T]_{d})$. The Bernoulli probabilities $\alpha_T: \mathcal{Z}\times\mathcal{K} \rightarrow (0, 1)^D$ are predicted by a neural component. We use a multilayer perceptron (MLP) $f_\theta$ which predicts the probability $p_\mathrm{flip}$ of binary variables in $z_0$ to toggle $p_\mathrm{flip} = f_\theta(z_0, k)$. 
The index operator $[x]_{d}$ returns the d\textsuperscript{th} element of vector $x$.

\paragraph{Objective}
For any state $s_0 \in \mathcal{S}$ with associated symbolic state $z_{0} = \Phi(s_0)$ we aim to learn $K$ skills $\pi(a \given s_t, k)$ which maximize the diversity in the set of reachable successor states $\{z^k_T = \Phi(\operatorname{apply}(E, \pi, s_0, k)) \:|\: k \in \mathcal{K}\}$. Jointly, we aim to model the effect of skill execution with the forward model $q_\theta(z_{T} \given z_{0}, k)$.
Inspired by Variational Intrinsic Control~\cite{gregor2017_vic} we take an information-theoretic perspective and maximize the mutual information $\mi(z_T, k \given z_0)$ between the skill index~$k$ and the symbolic observation~$z_T$ at skill termination given the symbolic observation~$z_0$ at skill initiation, i.e.,
    $\max \: \mi(z_T, k \given z_0) = \max H(z_T \given z_0) - H(z_T \given z_0, k)$.
The intuition behind this objective function is that we encourage the agent to (i) reach a diverse set of terminal observations  $z_T$ from an initial observation $z_0$ (by maximizing the conditional entropy $H(z_T \given z_0)$) and (ii) behave predictably such that the terminal observation $z_T$ is ideally fully determined by the initial observation $z_0$ and skill index $k$ (by minimizing $H(z_T \given z_0, k)$).
We reformulate the objective as an expectation over tuples $(s_0, k, s_T)$ by employing the mapping function~$\Phi$ as
$\mi(z_T, k \given z_0) = 
    \exn_{
        \substack{
            (s_0, k, s_T) \sim P
            }
        }
        \left[ 
            \log \frac{p(z_T \given z_0, k) }{ p(z_T \given z_0) }  
        \right]$
with $z_T := \Phi(s_T), z_0 := \Phi(s_0)$ and replay buffer $P$.
Similar to \cite{sharma2020dynamics} we derive a lower bound on the mutual information, which is maximized through the interplay of a RL problem and maximum likelihood estimation. 
To this end, we first introduce a variational approximation $q_\theta(z_T \given z_0, k)$ to the transition probability $p(z_T \given z_0, k)$, which we model by a neural component. 
We decompose 
\begin{equation}
    \label{eq:mi_decomposition}
    \mi(z_T, k \given z_0)  = 
    \exn_{\substack{(s_0, k, s_T) \sim P
    }} \left[ \log \frac{q_\theta(z_T \given z_0, k)}{ p(z_T \given z_0) }  \right] +\nonumber 
    \underbrace{\exn_{\substack{(s_0, k, s_T) \sim P  
    }} \left[ \log \frac{p(z_T \given z_0, k)}{q_\theta(z_T \given z_0, k) )} \right]}_{
    \approx \mathrm{D}_\mathrm{KL} ( p(z_T \given z_0, k) \:||\: q_\theta(z_T \given z_0, k) )}
\end{equation}
giving rise to the lower bound
$
    \mi(z_T, k \given z_0) \geq \exn_{\substack{(s_0, k, s_T) \sim P
    }} \left[ \log \frac{q_\theta(z_T \given z_0, k)}{ p(z_T \given z_0) }  \right]
$
whose maximization can be interpreted as a sparse-reward RL problem with reward 
$
    \hat{R}_T(k) = \log \frac{q_\theta(z_T \given z_0, k)}{ p(z_T \given z_0) }
$.
We approximate $p(z_T \given z_0)$ as $p(z_T \given z_0) \approx \sum_{k'} q_\theta(z_T \given z_0, k') p(k' \given z_0) $ and assume $k$ uniformly distributed and independent of $z_0$, i.e. $p(k' \given z_0) = \frac{1}{K}$. This yields a tractable reward
\begin{equation}
    \label{eq:reward}
    R_T(k) = \log \frac{q_\theta(z_T \given z_0, k)}{ \sum_{k'} q_\theta(z_T \given z_0, k') } + \log K.
\end{equation}
In Sec.~\ref{sec:reward_adjustments} we describe modifications we apply to the intrinsic reward $R_T$ which improve the performance of our proposed algorithm.
To tighten the lower bound, the KL divergence term in eq.~\eqref{eq:mi_decomposition} has to be minimized. Minimizing the KL divergence term corresponds to ``training'' the symbolic forward model $q_\theta$ by maximum likelihood estimation of the parameters $\theta$.

\paragraph{Training procedure}
In each epoch of training, we first collect skill trajectories on the environment using the skill policy~$\pi$. For each episode~$i \in \{1, ... , N\}$ we reset the environment and obtain an initial state $s_0^i$. Next, we uniformly sample skills~$k^i \sim \mathrm{Uniform}\{1, ..., K\}$. By iteratively applying the skill policy $\pi(\cdot |\cdot, {k^i})$ we obtain resulting states $s_0^i ... s_{T_i}^i$ and actions $a_0^i ... a_{T_i-1}^i$. A skill rollout terminates either if an environment-dependent step-limit is reached or when a change in the symbolic observation $z_t \neq z_0$ is observed. We append the rollouts to a limited-size buffer $\mathcal{B}$. In each training epoch we sample two sets of episodes $\mathcal{S}_\mathrm{RL}, \mathcal{S}_\mathrm{FM}$ from the buffer for training the policy $\pi$ and symbolic forward model $q_\theta$. Both episode sets are relabelled as described in Sec.~\ref{sec:relabelling}.
Let $i \in \{1, ... , M\}$ now refer to the episode index in the set $\mathcal{S}_\mathrm{RL}$. From $\mathcal{S}_\mathrm{RL}$ we sample transition tuples $([s^i_t, k^i], [s^i_{t+1}, k^i], a^i_t, r^i_{t+1}(k^i))$ which are used to update the policy $\pi$ using the soft actor-critic (SAC) algorithm \cite{haarnoja2018soft}. To condition the policy on skill~$k$ we concatenate $k$ to the state~$s$ as denoted by $[\cdot, \cdot]$. We set the intrinsic reward to zero $r^i_{t+1} = 0$ except for the last transition in an episode ($t+1 = T_i$) in which $r^i_{t+1}(k^i) = R(k^i)$. From the episodes in $\mathcal{S}_\mathrm{FM}$ we form tuples $(z^i_0 = \Phi(s^i_0), k^i, z^i_{T_i} = \Phi(s^i_{T_i}))$ which are used to train the symbolic forward model using gradient descent.

\paragraph{Relabelling}
\label{sec:relabelling}
Early in training, the symbolic transitions caused by skill executions mismatch the predictions of the symbolic forward model. We can in \emph{hindsight} increase the match between skill transitions and forward model by replacing the actual $k^i$ which was used to collect the episode $i$ by a different $k_*^i$ in all transition tuples $([s^i_t, k_*^i], [s^i_{t+1}, k_*^i], a^i_t, r^i_{t+1}(k_*^i))$ and $(z^i_0, k_*^i, z^i_{T_i})$ of episode $i$. In particular, we aim to replace $k^i$ by $k_*^i$ which has highest probability $k_*^i = \max_k q_\theta(k \given z_T, z_0)$. However, this may lead to an unbalanced distribution over $k_*^i$ after relabelling which is no longer uniform. To this end, we introduce a constrained relabelling scheme as follows. We consider a set of episodes indexed by $i \in \{1,...,N\}$ and compute skill log-probabilities for each episode which we denote by $Q^i_{k} = \log q_\theta(k \given z_0^i, z_{T_i}^i)$ where 
    $q_\theta(k \given z_0^i, z_{T_i}^i) = \frac{q_\theta(z_{T_i}^i \given z_0^i,k)}{\sum_{{k'}} q_\theta(z_{T_i}^i \given z_0^i, {k'})}$.
We find a relabeled skill for each episode $(k_*^{1}, ..., k_*^{N})$ which maximizes the scoring $\max_{(k_*^{1}, ..., k_*^{N})} \sum_i Q^i_{k_*^i}$
under the constraint that the counts of re-assigned skills $(k_*^{1}, ..., k_*^{N})$ and original skills $(k^{1}, ..., k^{N})$ match, i.e.
    $\#_{i=1}^N[k_*^i = k] = \#_{i=1}^N[k^i = k] \quad \forall k \in \{1,...,K\}$
which is to ensure that after relabelling no skill is over- or underrepresented. 
The count operator $\#[\cdot]$ counts the number of positive (true) evaluations of its argument in square brackets.
This problem can be formulated as a linear sum assignment problem which we solve using the Hungarian method 
\cite{kuhn1955hungarian, munkres1957algorithms}.
While we pass all episodes in $\mathcal{S}_\mathrm{FM}$ to the relabelling module, only a subset  ($50\%$) of episodes in $\mathcal{S}_\mathrm{RL}$ can potentially be relabeled to retain negative examples for the SAC agent. Relabelling experience in hindsight to improve sample efficiency is a common approach in goal-conditioned \cite{andrychowicz2017hindsight} and hierarchical \cite{levy2019_hac} RL.

\paragraph{Reward improvements}
\label{sec:reward_adjustments}
The reward in eq.~\eqref{eq:reward} can be denoted as 
$R(k) = \log q_\theta(k \given z_0, z_T) + \log K$. For numerical stability, we define a lower bounded term $\bar{Q}_k = \operatorname{clip}(\log q_\theta(k \given z_0, z_T), \mathrm{min}=-2\log(K))$ and write $R^0(k) = \bar{Q}_k + \log K$.
In our experiments, we observed that occasionally the agent is stuck in a local minimum in which (i) the learned skills are not unique, i.e., two or more skills $k \in \mathcal{K}$ cause the same symbolic transition $z_0 \rightarrow z_T$. In addition, (ii), occasionally, not all possible symbolic transitions are discovered by the agent. To tackle (i) we reinforce the policy $\pi$ with a positive reward if and only if no other skill $k'$ better fits the symbolic transition ($z_0 \rightarrow z_T$) generated by $\operatorname{apply}(E, \pi, s_0, k)$, i.e.,
    $R^\mathrm{norm}(k) = \bar{Q}_k - \mathrm{top2}_{k'} \bar{Q}_{k'}$.
which we call \textbf{second-best normalization}.
The operator $\mathrm{top2}_{k'}$ selects the second-highest value of its argument for $k' \in \mathcal{K}$. We define $R^\mathrm{base}(k) = R^\mathrm{norm}(k)$ except for the ``No second-best norm.'' ablation where $R^\mathrm{base}(k) = R^0(k)$.
To improve (ii) the agent obtains a \textbf{novelty bonus} for transitions ($z_0 \rightarrow z_T$) which are not modeled by the symbolic forward model for \emph{any} $k'$ by
    $R(k) = R^\mathrm{base}(k) - \max_{k'} \log q_\theta(z_T \given z_0, k')$.
If the symbolic state does not change ($z_T = z_0$), we set $R(k) = -2 \log(K)$ (minimal attainable reward).

\paragraph{Planning and skill execution}
\label{sec:planning}
A task is presented to our agent as an initial state of the environment $s_0$ with associated symbolic observation $z_0$ and a symbolic goal $z^*$. First, we leverage our learned symbolic forward model $q_\theta$ to plan a sequence of skills $k_1, ..., k_N$ from $z_0$ to $z^*$ using breadth-first search (BFS). We use the mode of the distribution over $z'$ for node expansion in BFS: $\operatorname{successor}_{q_\theta}(z, k) = \operatorname{argmax}_{z' \in \mathcal{Z}} q_\theta(z' \given z, k)$. After planning, the sequence of skills $[k_1, ..., k_N]$ is iteratively applied to the environment through $s_{(n)} = \operatorname{apply}(E, \pi, s_{(n-1)}, k_n)$. Inaccuracies of skill execution (leading to different symbolic observations than predicted) can be coped with by replanning after each skill execution. Both single-outcome (mode) determinisation and replanning are common approaches to probabilistic planning \cite{yoon2007ffreplan}. 
We provide further details in the supplementary material.

\section{Experiments}
\begin{figure}
  \centering
  \includegraphics[width=0.99\linewidth]{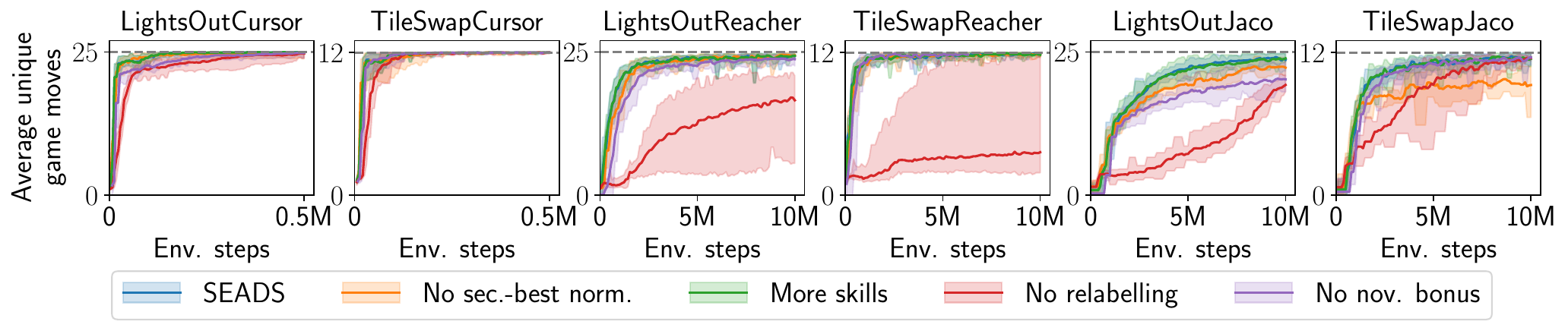} \\
  \includegraphics[width=0.99\linewidth]{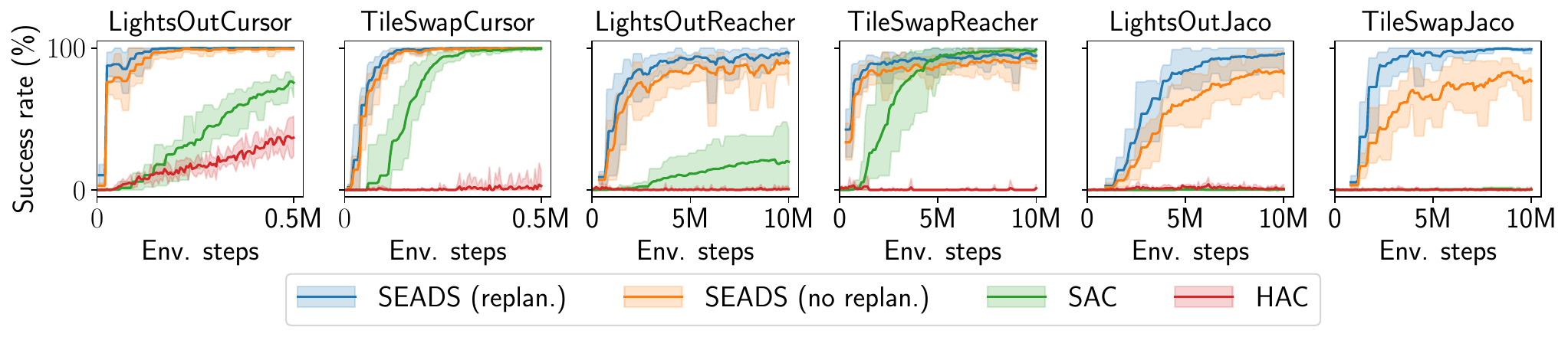}
  \caption{Top row: Number of learned unique game moves for ablations of \agent. Bottom row: Success rate of the proposed SEADS agent and baseline methods on \LightsOut, \TileSwap games embedded in \Cursor, \Reacher, \Jaco environments.  \agent performs comparably or outperforms the baselines on all tasks. The solid line depicts the mean, shaded area min. and max. of 10 (\agent, SAC on \Cursor) / 5 (HAC, SAC on \Reacher, \Jaco) independently trained agents.}
  \label{fig:performance_skillquality}
\end{figure}

We evaluate our proposed agent on a set of physically-embedded game environments. We follow ideas from \cite{mirza2020_physembplan} but consider single-player games which in principle enable full control over the environment without the existence of an opponent. We chose \emph{LightsOut} and \emph{TileSwap} as board games which are embedded in a physical manipulation scenario with \emph{Cursor}, \emph{Reacher} or \emph{Jaco} manipulators (see Fig.~\ref{fig:embedded_environments}).
The \LightsOut game (see Figure~\ref{fig:embedded_environments}) consists of a $5 \times 5$ board of fields. 
Each field has a binary illumination state of \emph{on} or \emph{off}. 
By pushing a field, its illumination state and the state of the (non-diagonally) adjacent fields toggles. 
At the beginning of the game, the player is presented a board where some fields are on and the others are off.
The task of the player is to determine a set of fields to push to obtain a board where all fields are off. The symbolic observation in all \LightsOut environments represents the illumination state of all 25 fields on the board $\mathcal{Z} = \{0, 1\}^{5\times5}$.
In \TileSwap (see Fig.~\ref{fig:embedded_environments}) a $3 \times 3$ board is covered by chips numbered from 0 to 8 (each field contains exactly one chip). Initially, the chips are randomly assigned to fields. Two chips can be swapped if they are placed on (non-diagonally) adjacent fields. The game is successfully finished after a number of swap operations if the chips are placed on the board in ascending order. In all \TileSwap environments the symbolic observation represents whether the i-th chip is located on the j-th field $\mathcal{Z} = \{0, 1\}^{9\times9}$.
To ensure feasibility, we apply a number of random moves (pushes/swaps) to the goal board configuration of the respective game. We quantify the difficulty of a particular board configuration by the number of moves required to solve the game. We ensure disjointness of board configurations used for training and testing through a hashing algorithm (see supp. material).
A board game move (``push'' in \LightsOut , ``swap'' in \TileSwap) is triggered by the manipulator's end effector touching a particular position on the board.
We use three manipulators of different complexity (see Fig.~\ref{fig:embedded_environments}).
%
The \emph{Cursor} manipulator can be navigated on the 2D game board by commanding $x$ and $y$ displacements. The board coordinates are $x, y \in [0, 1]$, the maximum displacement per timestep is $\Delta x, \Delta y = 0.2$. A third action triggers a push (\LightsOut) or swap (\TileSwap) at the current position of the cursor.
The \Reacher~\cite{tassa2018dmcontrolsuite} manipulator consists of a two-link arm with two rotary joints. The position of the end effector in the 2D plane can be controlled by external torques applied to the two rotary joints. As for the \emph{Cursor} manipulator an additional action triggers a game move at the current end effector coordinates.
The \Jaco manipulator \cite{campeaulecours2017kinova} is a 9-DoF robotic arm whose joints are velocity-controlled at 10Hz. It has an end-effector with three ``fingers'' which can touch the underlying board to trigger game moves. The arm is reset to a random configuration above the board around the board's center after a game move (details in supplementary material). 
By combining the games of \LightsOut and \TileSwap with the \Cursor, \Reacher and \Jaco manipulators we obtain six environments. 
For the step limit for skill execution we set $10$ steps on \Cursor and $50$ steps in \Reacher and \Jaco environments.
With our experiments we aim at answering the following research questions:
\textbf{R1:} How many distinct skills are learned by \agent? Does \agent learn all 25 (12) possible moves in \LightsOut (\TileSwap) reliably? 
\textbf{R2:} How do our design choices contribute to the performance of \agent?
\textbf{R3:} How well does \agent perform in solving the posed tasks in comparison to other flat and hierarchical RL approaches?
\textbf{R4:} Can our \agent also be trained and applied on a real robot?

\paragraph{Skill learning evaluation}
\label{sec:skill_discovery}
To address \textbf{R1} we investigate how many distinct skills are learned by \agent. If not all possible moves within the board games are learned as skills (25 for \LightsOut, 12 for \TileSwap), some initial configurations can become unsolvable for the agent, negatively impacting task performance. To count the number of learned skills we apply each skill $k \in \{1,...,K\}$ on a fixed initial state $s_0$ of the environment $E$ until termination (i.e., $\operatorname{apply}(E, s_0, \pi, k)$). Among these $K$ skill executions we count the number of unique game moves being triggered. We report the average number of unique game moves for $N=100$ distinct initial states $s_0$. 
On the \Cursor environments \agent detects nearly all possible game moves (avg. approx. 24.9 of 25 possible in \LightsOutCursor, 12 of 12 in \TileSwapCursor). For \Reacher almost all moves are found (24.3/11.8), while in the \Jaco environments some moves are missing occasionally (23.6/11.5), for the last checkpoints taken before reaching $5 \cdot 10^5$ (\Cursor), $10^7$ (\Reacher, \Jaco) env. steps (see Sec.~\ref{sec:seads:appendix_checkpointing} for details).
We demonstrate superior performance compared to a baseline skill discovery method (Variational Intrinsic Control, \citet{gregor2017_vic}) in the supplementary material.
We substantiate our agent design decisions through an ablation study (\textbf{R2}) in which we compare the number of unique skills (game moves) detected for several variants of \agent (see Fig.~\ref{fig:performance_skillquality}).
In a first study, we remove single parts from our agent to quantify their impact on performance. 
This includes training \agent without the proposed \emph{relabelling}, \emph{second-best normalization} and \emph{novelty bonus}. 
We found all of these innovations to be important for the performance of \agent, with the difference to the full \agent agent being most prominent in the \LightsOutJaco environment. Learning with \emph{more skills}  (15 for \TileSwap, 30 for \LightsOut) than actually needed does not harm performance.

\paragraph{Task performance evaluation}
\label{sec:performance_evaluation}
To evaluate the task performance of our agent and baseline agents (\textbf{R3}) we initialize the environments such that the underlying board game requires at maximum 5 moves (pushes in \LightsOut, swaps in \TileSwap, limited due to branching factor and BFS) to be solved. 
We evaluate each agent on 20 examples for each number of moves in $\{1, ..., 5\}$ required to solve the game.
We consider a task to be successfully solved if the target board configuration was reached (all fields \emph{off} in \LightsOut, ordered field in \TileSwap).
For \agent we additionally count tasks as ``failed'' if planning exceeds a wall time limit of $60$ seconds. We evaluate both planning variants with and without replanning.
%
As an instance of a flat (non-hierarchical) agent we evaluate the performance of Soft Actor-Critic (SAC~\cite{haarnoja2018soft}). 
The SAC agent receives the full environment state $s \in \mathcal{S}$ which includes the symbolic observation (board state). It obtains a reward of $1$ if it successfully solved the game and $0$ otherwise.
%
%
\begin{wrapfigure}{r}{0.25\textwidth}
  \begin{center}
    \vspace{-10pt}
    \includegraphics[width=0.25\textwidth]{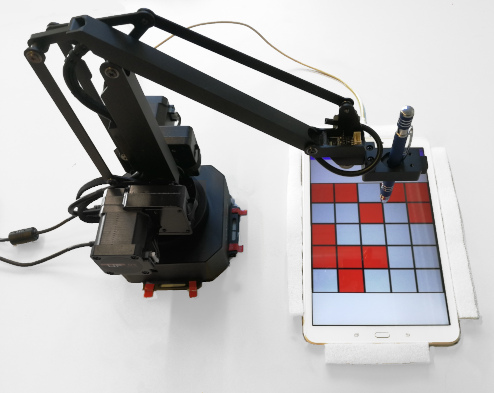}
    \vspace{-10pt}
  \end{center}
  \caption{Robot setup. }
  \label{fig:robot}
  \vspace{-10pt}
\end{wrapfigure}
In contrast to the Soft Actor-Critic agent the \agent agent leverages the decomposition of state $s \in \mathcal{S}$ and symbolic observation $z \in \mathcal{Z}$.
For a fair comparison to a hierarchical agent, we consider Hierarchical Actor-Critic (HAC,~\citet{levy2019_hac}), which, similar to \agent, can also leverage the decomposition of $s$ and $z$. 
We employ a two-level hierarchy in which the high-level policy sets \emph{symbolic} subgoals $z \in \mathcal{Z}$ to the low-level policy, thereby leveraging the access to the symbolic observation. We refer to the supplementary material for implementation and training details of SAC and HAC.
Fig.~\ref{fig:performance_skillquality} visualizes the performance of \agent and the baselines. On all environments \agent performs similar or outperforms the baselines, with the performance difference being most pronounced on the \Jaco environments on which SAC and HAC do not make any progress. 
On the  \Cursor environments \agent achieves a success rate of 100\%. On the remaining environments, the average success rate (with replanning) is 95.8\% (\LightsOutReacher), 95.5\% (\TileSwapReacher), 94.9\% (\LightsOutJaco), 98.8\% (\TileSwapJaco). These results are obtained at the last common checkpoints before reaching $5\cdot10^5$ (\Cursor) / $10^7$ (\Reacher, \Jaco) env. steps (see Sec.~\ref{sec:seads:appendix_checkpointing} for details).

\paragraph{Robot experiment}
\label{sec:robot}
To evaluate the applicability of our approach on a real-world robotic system (\textbf{R4}) we set up a testbed with a uArm Swift Pro robotic arm which interacts with a tablet using a capacitive pen (see Fig.~\ref{fig:robot}). 
The \agent agent commands a displacement $|\Delta x|, |\Delta y| \leq 0.2$ and an optional pushing command as in the \Cursor environments.
The board state is communicated to the agent through the tablet's USB interface. We manually reset the board \emph{once} at the beginning of training, and do not interfere in the further process. 
After training for $\approx 160k$ interactions ($\approx 43.5$ hours) the agent successfully solves all boards in a test set of 25 board configurations (5 per solution depth in $\{1,...,5\}$). We refer to the supplementary material for details and a video.

\vspace{-0.5em}
\section{Assumptions and Limitations}
\vspace{-0.5em}
Our approach assumes that the state abstraction is known, the symbolic observation $z$ is provided by the environment, and that the continuous state is fully observable.
Learning the state abstraction too is an interesting direction for future research.
The breadth-first search planner we use for planning on the symbolic level exhibits scaling issues for large solution depths; e.g., for LightsOut it exceeds a 5-minute threshold for solution depths (number of initial board perturbations) $\geq$ 9. In future work, more efficient heuristic or probabilistic planners could be explored. Currently, our BFS planner produces plans which are optimal with respect to the number of skills executed. Means for predicting and taking the skill execution cost into account for planning could be pursued in future work. In the more complex environments (\Reacher, \Jaco) we observe our agent to not learn all possible skills reliably, in particular for skills for which no transitions exist in the replay buffer. 
In future work one could integrate additional exploration objectives which incentivize to visit unseen regions of the state space.
Also, the approach is currently limited to settings such as in board games, where all symbolic state transitions should be mapped to skills.
It is an open research question how our skill exploration objective could be combined with demonstrations or task-specific objectives to guide the search for symbolic actions and limit the search space in more complex environments.
\vspace{-0.5em}
\section{Conclusion}
\vspace{-0.5em}
We present an agent which, in an unsupervised way, learns diverse skills in complex physically embedded board game environments which relate to \emph{moves} in the particular games. 
We assume a state abstraction from continuous states to symbolic states known and observable to the agent as prior information, and that skills lead to changes in the symbolic state.
The jointly learned forward model captures the temporally extended \emph{effects} of skill execution. 
We leverage this forward model to plan over a sequence of skills (moves) to solve a particular task, i.e., bring the game board to a desired state. 
We demonstrate that with this formulation we can solve complex physically embedded games with high success rate, that our approach compares favorably with other flat and hierarchical RL algorithms, and also transfers to a real robot.
Our approach provides an unsupervised learning alternative to prescribing the action abstraction and pretraining each skill individually before learning a forward model from skill executions. 
In future research, our approach could be combined with state abstraction learning to leverage its full potential.

\clearpage
\acknowledgments{This work was supported by Max Planck Society and Cyber Valley. The authors thank the International Max Planck Research School for Intelligent Systems (IMPRS-IS) for supporting Jan Achterhold.}


\printbibliography

\clearpage
\appendix
\begin{refsection}

\renewcommand{\thesection}{S.\arabic{section}}
\renewcommand{\thefigure}{S.\arabic{figure}}
\renewcommand{\thetable}{S.\arabic{table}}

\section{Introduction}
In the following we provide supplementary details and analysis of our approach. We show visualisations of the learned skills in Sec.~\ref{sec:suppl_learned_skills} and provide an analysis for solution depths $>5$ in Sec.~\ref{sec:suppl_highsoldepth}. In Sec.~\ref{sec:suppl_morelightsout} we introduce additional embedded \LightsOut environments, featuring a 3D board and larger board sizes, to evaluate the exploration capabilities and limitations of our method. In Sec.~\ref{sec:suppl_robot_exp} we give details on the robot experiment, and provide additional ablation results in Sec.~\ref{sec:suppl_ablation_analysis}. 
Architectural and implementation details are given in Sec.~\ref{sec:suppl_details_seads} for \agent, in Sec.~\ref{sec:suppl_baseline_sac} for the SAC baseline and in Sec.~\ref{sec:suppl_baseline_hac} for the HAC baseline.
In Sec.~\ref{sec:suppl_baseline_vic} we compare \agent to a variant which uses a skill discriminator instead of a forward model as in "Variational Intrinsic Control" (VIC, \citet{gregor2017_vic}).
In Sec.~\ref{sec:suppl_env_details} we detail how we define training and test splits on our proposed environments.
Finally, we present detailed results for hyperparameter search on the SAC and HAC baselines in Sec.~\ref{sec:suppl_hyperparameter_search}.

\section{Learned skills}
\label{sec:suppl_learned_skills}
\begin{figure}[h]
\centering
\begin{subfigure}[t]{0.3\linewidth}
\centering
\includegraphics[width=\linewidth]{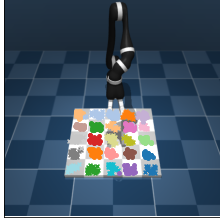}
\caption{Contact points of \Jaco end effector.}
\label{fig:skills_lightsout_jaco}
\end{subfigure}\hspace{1em}
\begin{subfigure}[t]{0.3\linewidth}
\centering
\includegraphics[width=\linewidth]{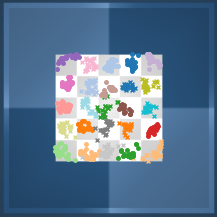}
\caption{Contact points of \Reacher end effector.}
\label{fig:skills_lightsout_reacher}
\end{subfigure}\hfill
\caption{Contact points of the \Jaco (\Reacher) end effector in the \LightsOutCursor (\LightsOutReacher) environments when executing skill $k \in \{1,...,25\}$ on 20 different initializations of the environment. Each skill is assigned a unique color/marker combination. We show the agent performance after $1\times10^7$ environment steps. We observe that the \agent agent learns to push individual fields as skills.}
\label{fig:learned_skills}
\end{figure}

We provide additional visualizations on the behaviour of \agent in Fig.~\ref{fig:learned_skills}, showing that \agent learns to assign skills to pushing individual fields on the game boards. 

\subsection{Skill trajectories}

\begin{figure}[tb]
\centering
\begin{subfigure}[t]{0.49\linewidth}
\centering
\includegraphics[width=\linewidth]{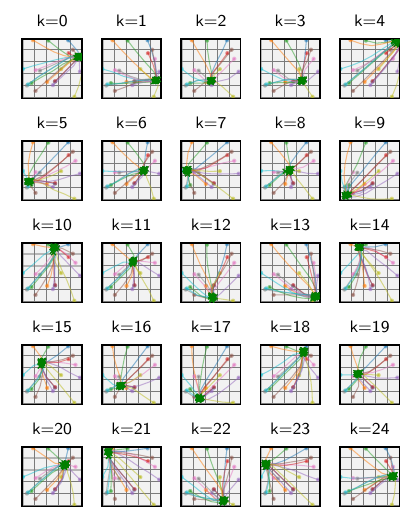}
\caption{Skill trajectories on \LightsOutCursor.}
\label{fig:skill_trajs_lightsout_cursor}
\end{subfigure}\hfill
\begin{subfigure}[t]{0.49\linewidth}
\centering
\includegraphics[width=\linewidth]{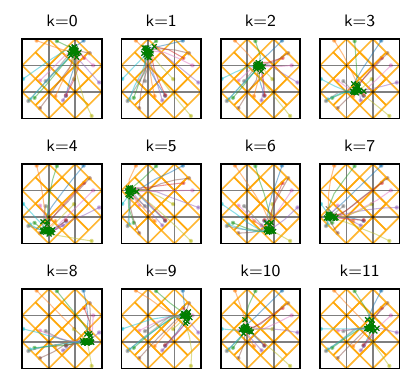}
\caption{Skill trajectories on \TileSwapCursor.}
\label{fig:skill_trajs_tileswap_cursor}
\end{subfigure}
\caption{Trajectories in \Cursor-embedded environments for skills $k$ on 20 different environment initializations. Colored lines show the $x,y$-coordinates of the Cursor, with the circular marker indicating the start position of the skill. Black markers indicate locations where a "push" was executed, and green markers where the push has caused a change in the symbolic state.}
\label{fig:learned_skills_cursor}
\end{figure}

\begin{figure}[tb]
\centering
\begin{subfigure}[t]{0.49\linewidth}
\centering
\includegraphics[width=\linewidth]{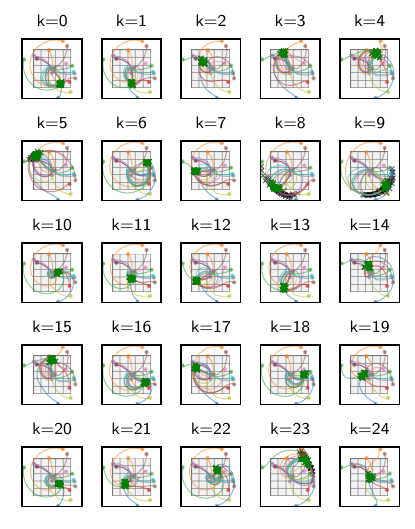}
\caption{Skill trajectories on \LightsOutReacher.}
\label{fig:skill_trajs_lightsout_reacher}
\end{subfigure}\hfill
\begin{subfigure}[t]{0.49\linewidth}
\centering
\includegraphics[width=\linewidth]{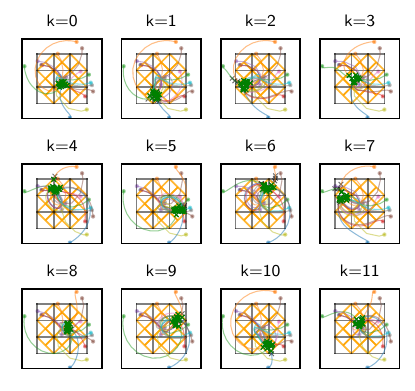}
\caption{Skill trajectories on \TileSwapReacher.}
\label{fig:skill_trajs_tileswap_reacher}
\end{subfigure}
\caption{Trajectories in \Reacher-embedded environments for skills $k$ on 20 different environment initializations. Colored lines show the $x,y$-coordinates of the Reacher end-effector, with the circular marker indicating the start position of the skill. Black markers indicate locations where a "push" was executed, and green markers where the push has caused a change in the symbolic state.}
\label{fig:learned_skills_reacher}
\end{figure}

\begin{figure}[tb]
\centering
\begin{subfigure}[t]{0.49\linewidth}
\centering
\includegraphics[width=\linewidth]{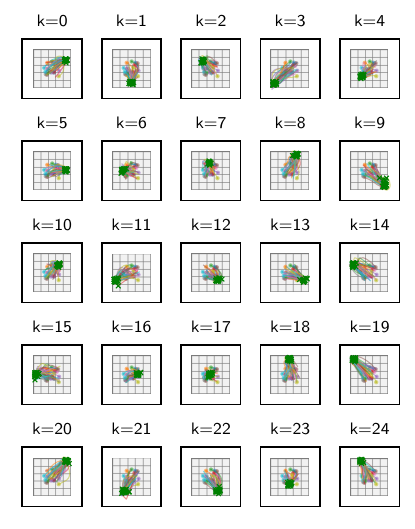}
\caption{Skill trajectories on \LightsOutJaco.}
\label{fig:skill_trajs_lightsout_jaco}
\end{subfigure}\hfill
\begin{subfigure}[t]{0.49\linewidth}
\centering
\includegraphics[width=\linewidth]{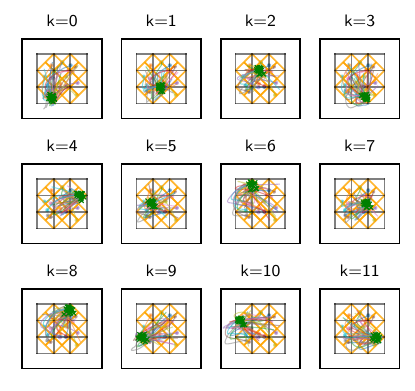}
\caption{Skill trajectories on \TileSwapJaco.}
\label{fig:skill_trajs_tileswap_jaco}
\end{subfigure}
\caption{Trajectories in \Jaco-embedded environments for skills $k$ on 20 different environment initializations. Colored lines show the $x,y$-coordinates of the Jaco hand, with the circular marker indicating the start position of the skill. Green markers indicate contact locations with the board.}
\label{fig:learned_skills_jaco}
\end{figure}

In Figs.~\ref{fig:learned_skills_cursor},~\ref{fig:learned_skills_reacher} and~\ref{fig:learned_skills_jaco} we provide visualizations of trajectories executed by the learned skills on the \Cursor, \Reacher and \Jaco environments.

\subsection{Solution length}
\label{sec:suppl_solution_length}

\begin{figure}[tb]
  \centering
  \includegraphics[width=0.99\linewidth]{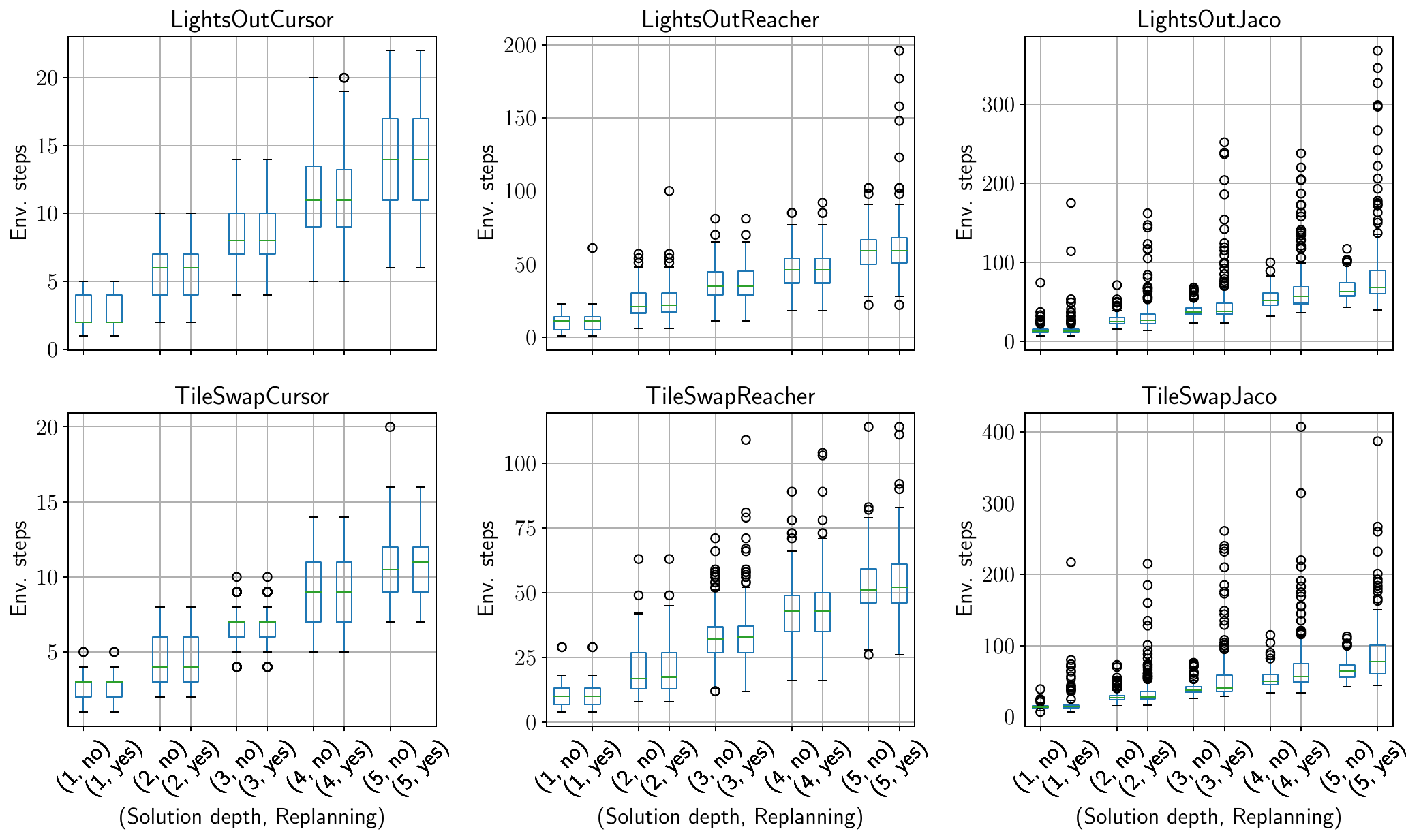} \\
  \caption{Analysis of solution lengths (total number of manipulator steps required to solve the physically embedded board games) for different environments, solution depths and planning with/without re-planning. Box-plots show the 25\%/75\% quartiles and median (green line). Whiskers extend to the farthest datapoint within the 1.5-fold interquartile range. Outliers are plotted as circles.}
  \label{fig:suppl_solution_length}
\end{figure}

In Fig.~\ref{fig:suppl_solution_length} we provide an analysis how many low-level environment steps (i.e., manipulator actions) are executed to solve instances of the presented physically embedded board games. We follow the evaluation procedure of the main paper (Fig. 3) and show results for 10 trained agents and 20 initial board configurations for each solution depth in $\{1,...,5\}$. We only report results on board configurations which are successfully solved by \agent. We observe that even for a solution depth of 5, for the Cursor environments, only relatively few environment steps are required in total to solve the board game ($\approx 15$ for \LightsOutCursor, $\approx 10$ for \TileSwapCursor). In contrast, the more complex \Reacher and \Jaco environments require significantly more interactions to be solved, with up to 400 steps executed on \TileSwapJaco for a solution depth of 5 and enabled re-planning.

\subsection{Skill length}
\begin{figure}[h!]
  \centering
  \includegraphics[width=0.5\linewidth]{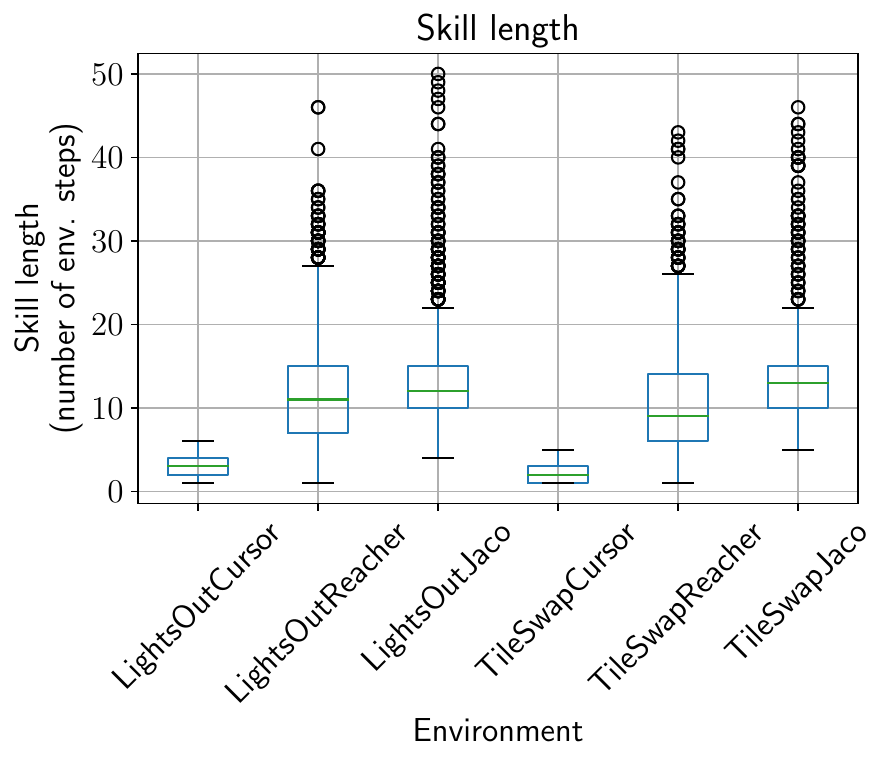} \\
  \caption{Analysis of skill lengths (number of manipulator steps executed within a single skill) for different environments. Box-plots show the 25\%/75\% quartiles and median (green line). Whiskers extend to the farthest datapoint within the 1.5-fold interquartile range. Outliers are plotted as circles.}
  \label{fig:suppl_skill_length}
\end{figure}

Following the evaluation procedure from Sec.~\ref{sec:suppl_solution_length} we report the distribution of skill lengths (i.e., number of actions applied to the manipulator per skill) in Fig.~\ref{fig:suppl_skill_length}. Again, we only include problem instances which were successfully solved by \agent. While skill executions on the \Cursor environments are typically short (median 3/2 manipulator actions for \LightsOutCursor/\TileSwapCursor), the \Reacher and \Jaco environments require a higher number of manipulator actions per skill (median 11/12/9/13 for \LightsOutReacher/\LightsOutJaco/\TileSwapReacher/\TileSwapJaco).

\clearpage
\section{Large solution depth analysis}
\label{sec:suppl_highsoldepth}
\begin{figure}[h!]
\centering
\includegraphics[height=10em]{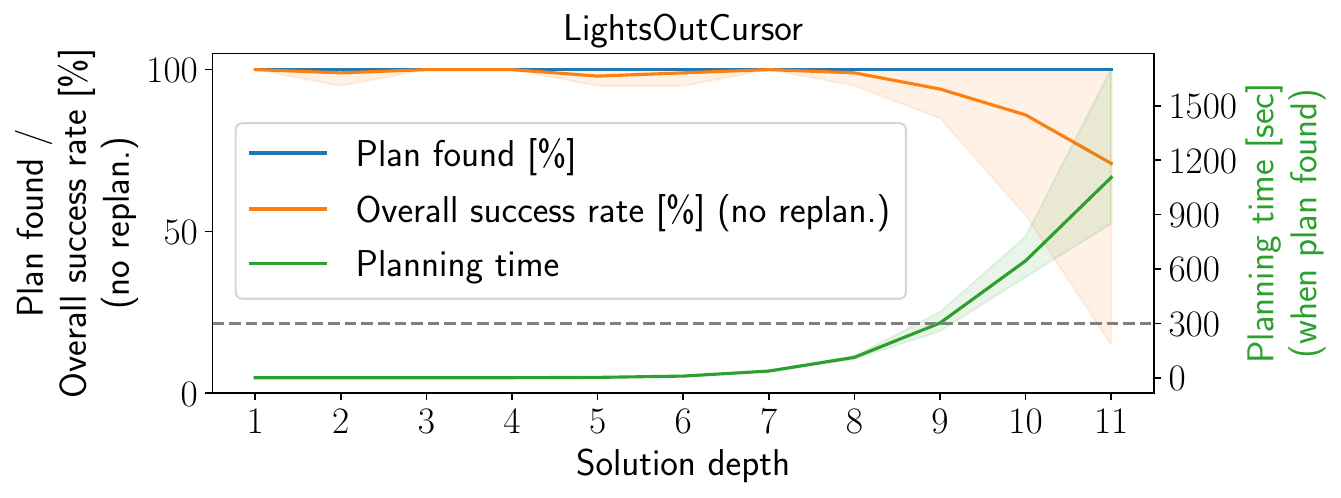}
\caption{Analysis of the \LightsOutCursor environment for solution depths $> 5$. "Solution depth" is the number of game moves required to solve the \LightsOut instance. "Plan found" refers to the ratio of problem instances where BFS finds a feasible plan. "Overall success rate" quantifies the ratio of problem instances for which a feasible plan was found and, in addition, was successfully executed by the low-level policy. The "Planning time" refers to the wall-time the breadth-first search planner runs, for the cases in which it finds a feasible plan. We report results for 20 problem instances for each solution depth for 5 independently trained agents. We refer to sec.~\ref{sec:suppl_highsoldepth} for details. The solid line refers to the mean, shaded area to min/max over 5 independently trained agents.} 
\label{fig:suppl_high_soldepth_lightsout}
\end{figure}

In Fig.~\ref{fig:suppl_high_soldepth_lightsout} we present an analysis for solving \LightsOut tasks with solution depths $> 5$ using the learned \agent agent on \LightsOutCursor. We observe a high mean success rate of $\geq 98\%$ for solution depths $\leq 8$. However, the time required for the breadth-first search planner to find a feasible plan increases from $\approx 2.02s$ for solution depth $5$ to $\approx 301.8s$ for solution depth $9$. We abort BFS planning if the list of nodes to expand exceeds a size of $\approx 16$ GB memory usage. All experiments were conducted on Intel\textregistered{}  Xeon\textregistered{} Gold 5220 CPUs with a clock rate of 2.20 GHz.

\section{Additional LightsOut variants}
\label{sec:suppl_morelightsout}
\subsection{LightsOut3DJaco}
\label{sec:suppl_lightsout3djaco}
\begin{figure}[h!]
\centering
\begin{subfigure}[t]{0.19\linewidth}
\includegraphics[width=\linewidth]{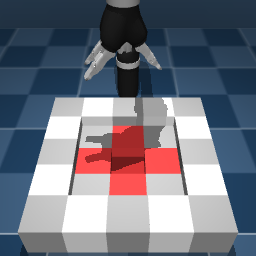}
\end{subfigure} \hfill
\begin{subfigure}[t]{0.19\linewidth}
\includegraphics[width=\linewidth]{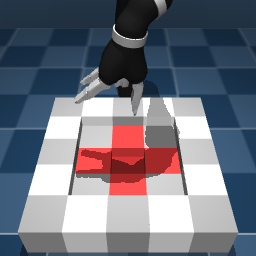}
\end{subfigure} \hfill
\begin{subfigure}[t]{0.19\linewidth}
\includegraphics[width=\linewidth]{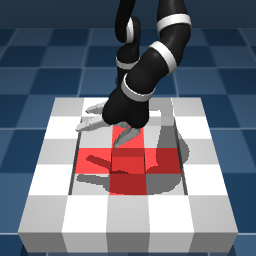}
\end{subfigure} \hfill
\begin{subfigure}[t]{0.19\linewidth}
\includegraphics[width=\linewidth]{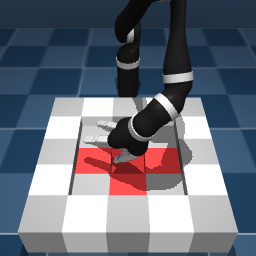}
\end{subfigure} \hfill
\begin{subfigure}[t]{0.19\linewidth}
\includegraphics[width=\linewidth]{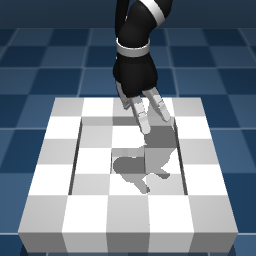}
\end{subfigure}
\caption{\texttt{LightsOut3DJaco}: A variant of the \LightsOutJaco environment with elevated fields. In comparison to the \LightsOutJaco environment this environment poses an additional challenge to the agent, as it has to avoid to push fields with its fingers accidentally during skill execution. We show the execution of a skill which has learned to push the center field for $T=\{0, 4, 8, 12, 14\}$.} 
\label{fig:suppl_lightsout3d}
\end{figure}

\begin{figure}[h!]
\centering
\begin{subfigure}[t]{0.3\linewidth}
\includegraphics[height=8em]{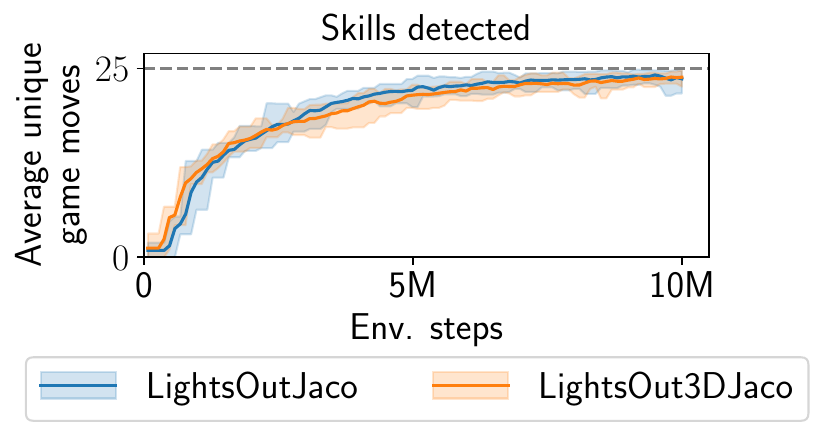}
\caption{Average number of unique game moves detected on \LightsOutJaco and \texttt{LightsOut3DJaco}.}
\label{fig:suppl_lightsout3djaco_skillcoverage}
\end{subfigure} \hfill
\begin{subfigure}[t]{0.6\linewidth}
\includegraphics[height=8em]{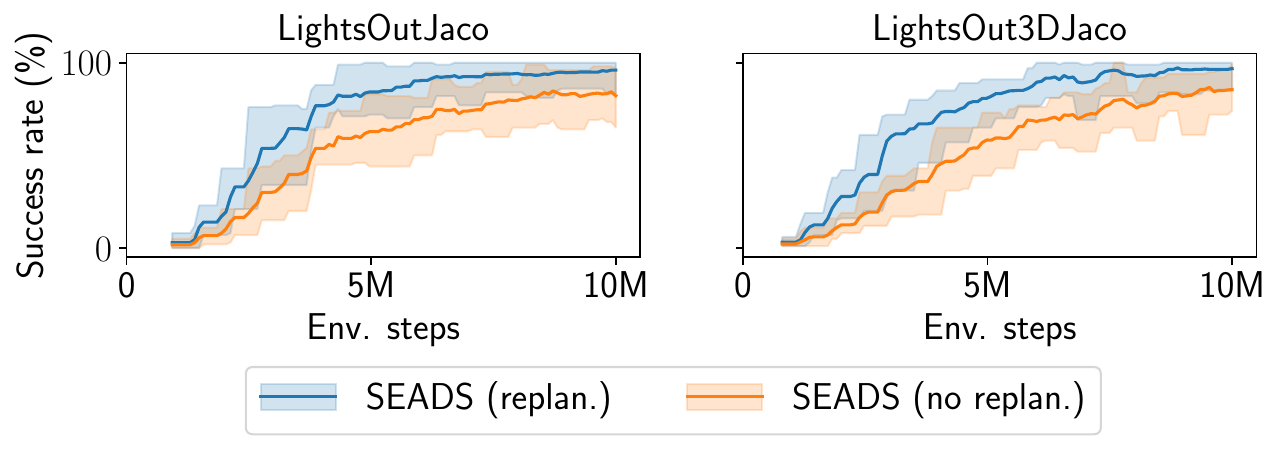}
\caption{\agent task performance on \LightsOutJaco and \texttt{LightsOut3DJaco}, with and without replanning.}
\label{fig:suppl_lightsout3djaco_successrate}
\end{subfigure}
\caption{Evaluation on number of (a) detected game moves  and (b) task performance  on \texttt{LightsOut3DJaco} (see sec.~\ref{sec:suppl_lightsout3djaco} for details).} 
\label{fig:suppl_lightsout3d_perf}
\end{figure}

In addition to the \LightsOutJaco environment presented in the main paper we introduce an additional environment \texttt{LightsOut3DJaco}. While in \LightsOutJaco the LightsOut board is a flat plane, in \texttt{LightsOut3DJaco} the fields are elevated/recessed depending on their distance to the board's center (see Fig.~\ref{fig:suppl_lightsout3d}). This poses an additional challenge to the agent, as it has to avoid to push fields with its fingers accidentally during skill execution.
Despite the increased complexity of \texttt{LightsOut3DJaco} over \LightsOutJaco we observe similar results in terms of detected game moves (Fig.~\ref{fig:suppl_lightsout3djaco_skillcoverage}) and task performance (Fig.~\ref{fig:suppl_lightsout3djaco_successrate}). Both environments can be solved with a high success rate of  $94.9\%$ (\texttt{LightsOutJaco}) and $96.3\%$ (\texttt{LightsOut3DJaco}). For the experiments, we follow the evaluation protocols from the main paper.

\subsection{Larger LightsOut boards / spacing between fields }
\begin{figure}[h!]
\centering
\begin{subfigure}[t]{0.19\linewidth}
\includegraphics[width=\linewidth]{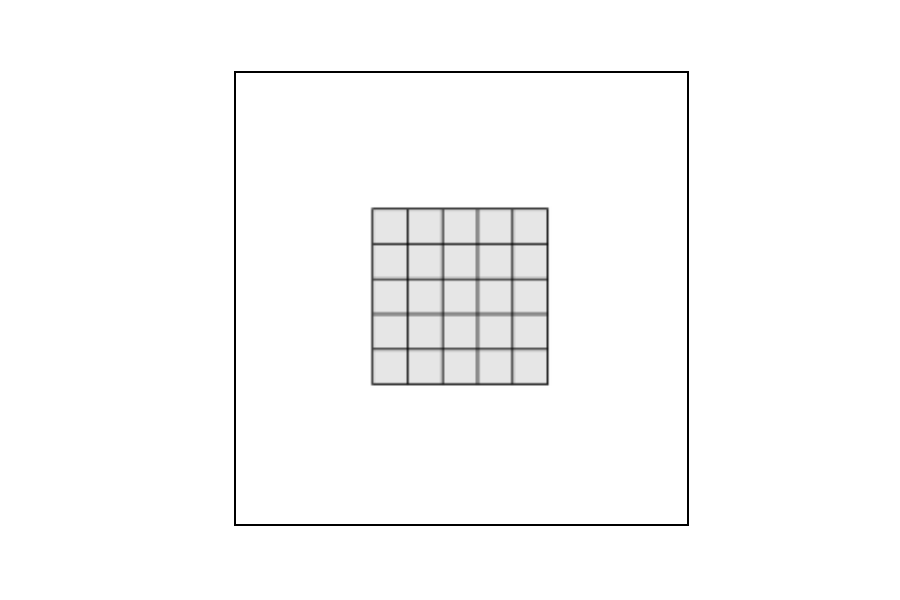}
\end{subfigure} \hfill
\begin{subfigure}[t]{0.19\linewidth}
\includegraphics[width=\linewidth]{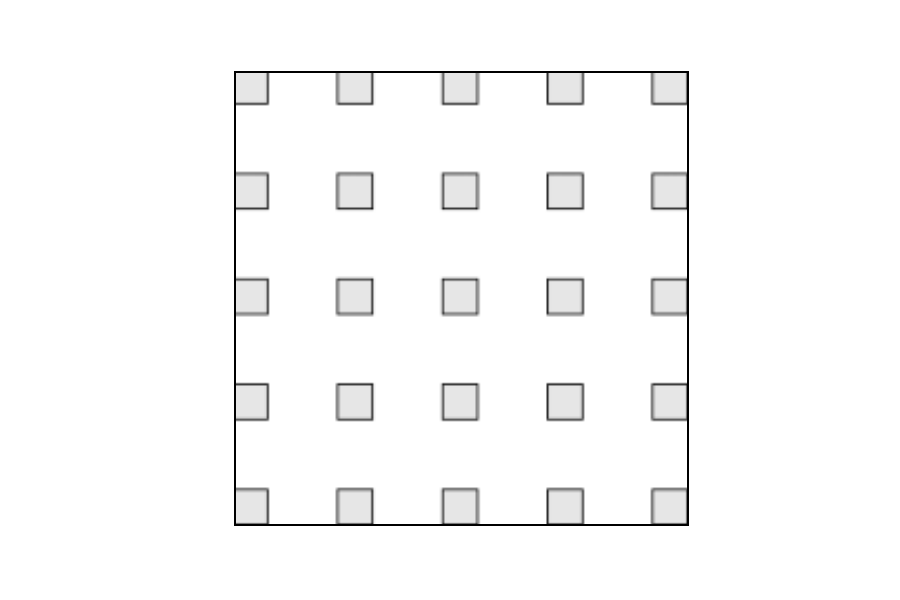}
\end{subfigure} \hfill
\begin{subfigure}[t]{0.19\linewidth}
\includegraphics[width=\linewidth]{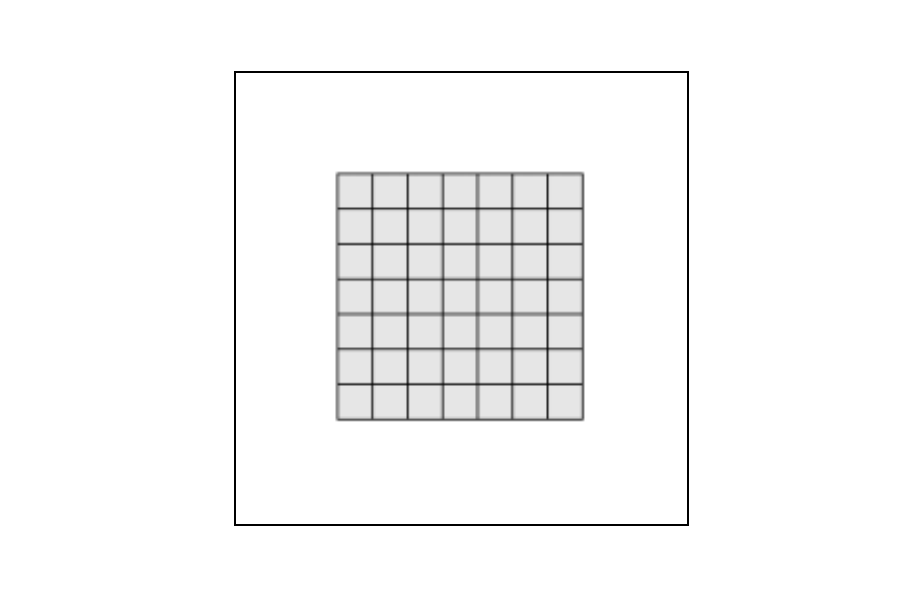}
\end{subfigure} \hfill
\begin{subfigure}[t]{0.19\linewidth}
\includegraphics[width=\linewidth]{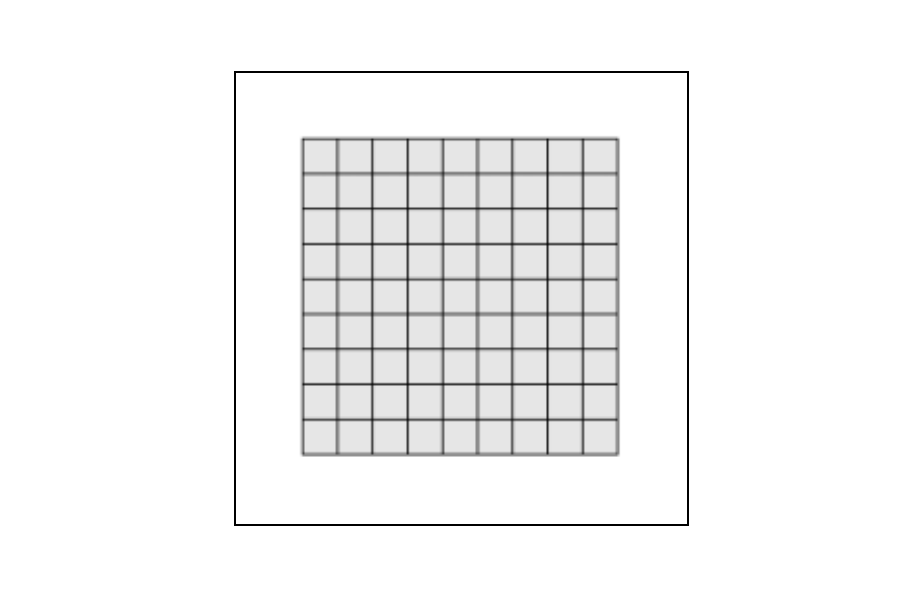}
\end{subfigure} \hfill
\begin{subfigure}[t]{0.19\linewidth}
\includegraphics[width=\linewidth]{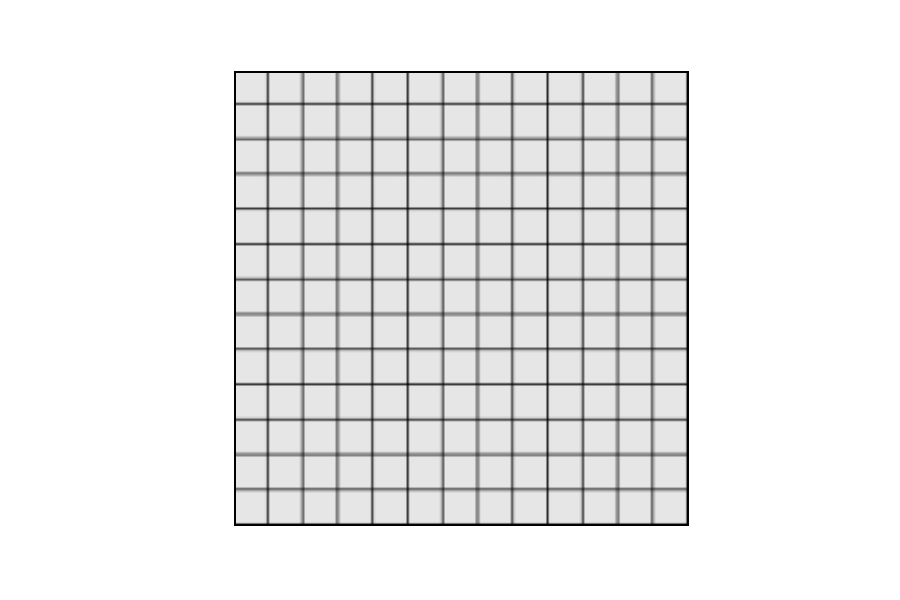}
\end{subfigure}
\caption{Variants of \LightsOutCursor for different board sizes ($5\times5$, $7\times7$, $9\times9$, $13\times13$) and spacing introduced between fields (second panel from left).}
\label{fig:suppl_lightsout_spaced}
\end{figure}

\begin{figure}[h!]
\centering
\begin{subfigure}[t]{0.265\linewidth}
\includegraphics[width=\linewidth]{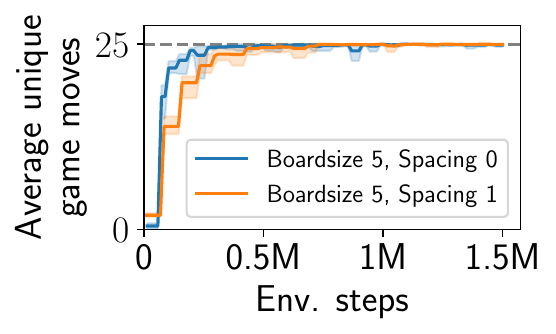}
\caption{Boardsize $5\times5$}
\label{fig:suppl_lightsout_spaced_results_5}
\end{subfigure} \hfill
\begin{subfigure}[t]{0.22\linewidth}
\includegraphics[width=\linewidth]{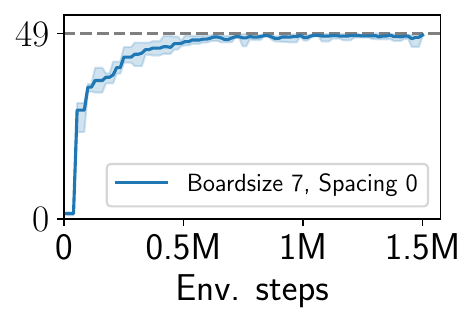}
\caption{Boardsize $7\times7$}
\end{subfigure} \hfill
\begin{subfigure}[t]{0.22\linewidth}
\includegraphics[width=\linewidth]{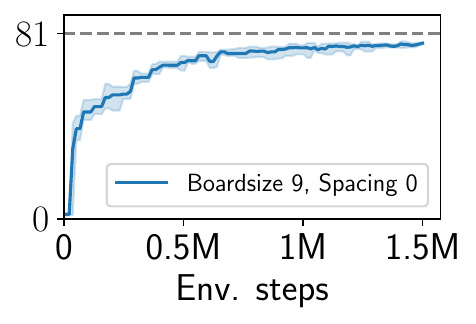}
\caption{Boardsize $9\times9$}
\end{subfigure} \hfill
\begin{subfigure}[t]{0.22\linewidth}
\includegraphics[width=\linewidth]{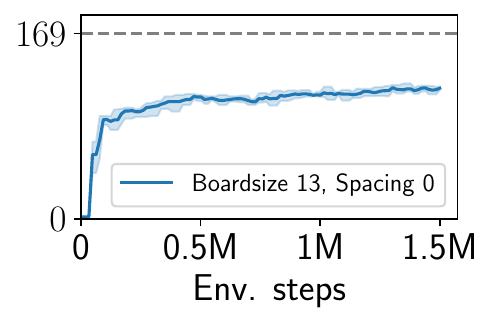}
\caption{Boardsize $13\times13$}
\end{subfigure} \hfill
\caption{Detected average unique game moves (skills) on \LightsOutCursor environment for different board sizes ($5\times5$, $7\times7$, $9\times9$, $13\times13$) and spacing. The introduced spacing slows down skill learning (a). While for boards until size $9 \times 9$ nearly all skills are found (a-c), the $13\times13$ environment poses a challenge to \agent (d). Solid line: mean / Shaded area: min/max over 5 independently trained agents.}
\label{fig:suppl_lightsout_spaced_results}
\end{figure}

{
In this experiment we investigate how well \agent performs on environments in which a very large number of skills has to be learned and where noisy executions of already learned skills uncover new skills with low probability. To this end, we modified the \LightsOutCursor environment to have more fields (and thereby, more skills to be learned), and introduced a spacing between the tiles, which makes detecting new skills more challenging. For a fair comparison, we keep the total actionable area in all environments constant, which introduces an empty area either around the board or around the tiles (see Fig.~\ref{fig:suppl_lightsout_spaced}). We make two main observations: (i) As presumed, learning skills in the \LightsOutCursor environment with spacing between tiles requires more environment interactions than for adjacent tiles (see Fig.~\ref{fig:suppl_lightsout_spaced_results_5}) (ii) For boards up to size $9 \times 9$ a large majority of skills is found after $1.5$ million environment steps of training ($5\times5: 24.9/25$ , $7\times7: 48.6/49$, $9\times9: 76.8/81$). The \LightsOutCursor environment with boardsize $13\times13$ poses a challenge to \agent with $119.3/169$ skills detected after $1.5$ million environment steps (see Fig.~\ref{fig:suppl_lightsout_spaced_results}). The numbers reported are averages over 5 independently trained agents.
}

\section{Real Robot Experiment}
\label{sec:suppl_robot_exp}

For the real robot experiment we use a uArm Swift Pro robotic arm that interacts with a \LightsOut board game. The board game runs on a Samsung Galaxy Tab A6 Android tablet with a screen size of 10.1 inches. 
We mapped the screen plane excluding the system status bar and action bar of the app (blue bar) to normalized coordinates $(x, y) \in [0, 1]$. A third $z \in [0, 1]$ coordinate measures the perpendicular distance to the screen plane, with $z=1$ approximately corresponding to a distance of \si{10\,\cm} to the screen. To control the robot arm, we use the Python SDK from \cite{uArm-Developer}, which allows to steer the end effector to $\overrightarrow{X} = (X, Y, Z)$ target locations in a coordinate frame relative to the robot's base. As the robot's base is not perfectly aligned with the tablet's surface, e.g. due to the rear camera, we employed a calibration procedure. We measured the location of the four screen corners in $(X,Y,Z)$ coordinates using the SDK's \texttt{get\_position} method (by placing the end effector holding the capacitive pen on the particular corners) and fitted a plane to these points minimizing the squared distance. We reproject the measured points onto the plane and compute a perspective transform by pairing the reprojected points with normalized coordinates $(x, y) \in \{0,1\}\times\{0,1\}$. To obtain robot coordinates $(X,Y,Z)$ from normalized coordinates $(x,y,z)$ we first apply the perspective transform on $(x,y)$, yielding $\hat{X} = (X, Y, Z=0)$. We subsequently add the plane's normal to $(X,Y,Z)$ scaled by $z$ and an additional factor which controls the distance to the tablet's surface for $z=1$. 
The state of the board is communicated to the host machine running \agent via USB through the logging functionality of the Android Debug Bridge. The whole system including robotic arm and Android tablet is interfaced as an \texttt{OpenAI Gym} \cite{brockman2016openaigym} environment. 

\subsection{Training on robot with absolute push position actions}
\label{sec:suppl_robot_exp_abspush}
\begin{figure}[h!]
\centering
\begin{subfigure}[t]{0.3\linewidth}
\centering
\includegraphics[width=\linewidth]{graphics/robot_setup_mod.jpg}
\caption{Robotic arm (uArm Swift Pro) interacting with a LightsOut game running on an Android tablet.}
\label{fig:suppl_robotexp_image}
\end{subfigure}\hspace{1em}
\begin{subfigure}[t]{0.3\linewidth}
\centering
\includegraphics[width=0.6\linewidth]{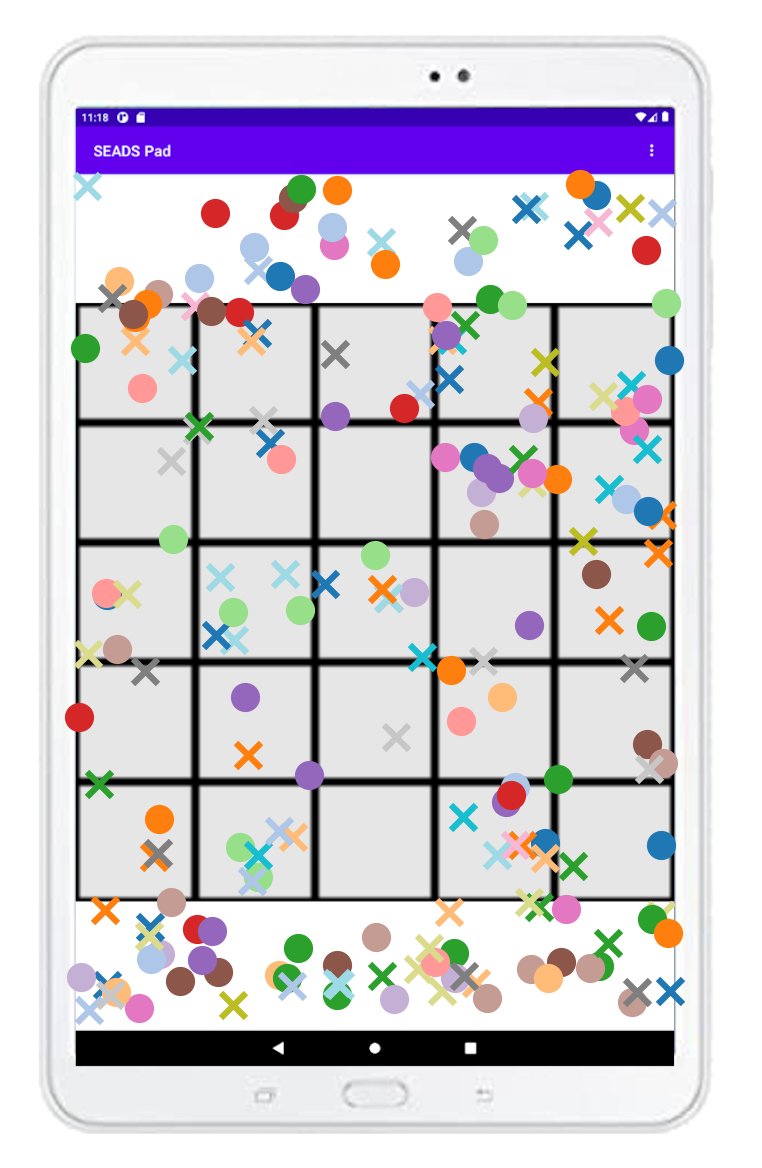}
\caption{Contact points of 200 skill executions with randomly sampled skills $k \in \{1,...,25\}$ at the beginning of training.}
\label{fig:suppl_robotexp_beginning}
\end{subfigure}\hspace{1em}
\begin{subfigure}[t]{0.3\linewidth}
\centering
\includegraphics[width=0.6\linewidth]{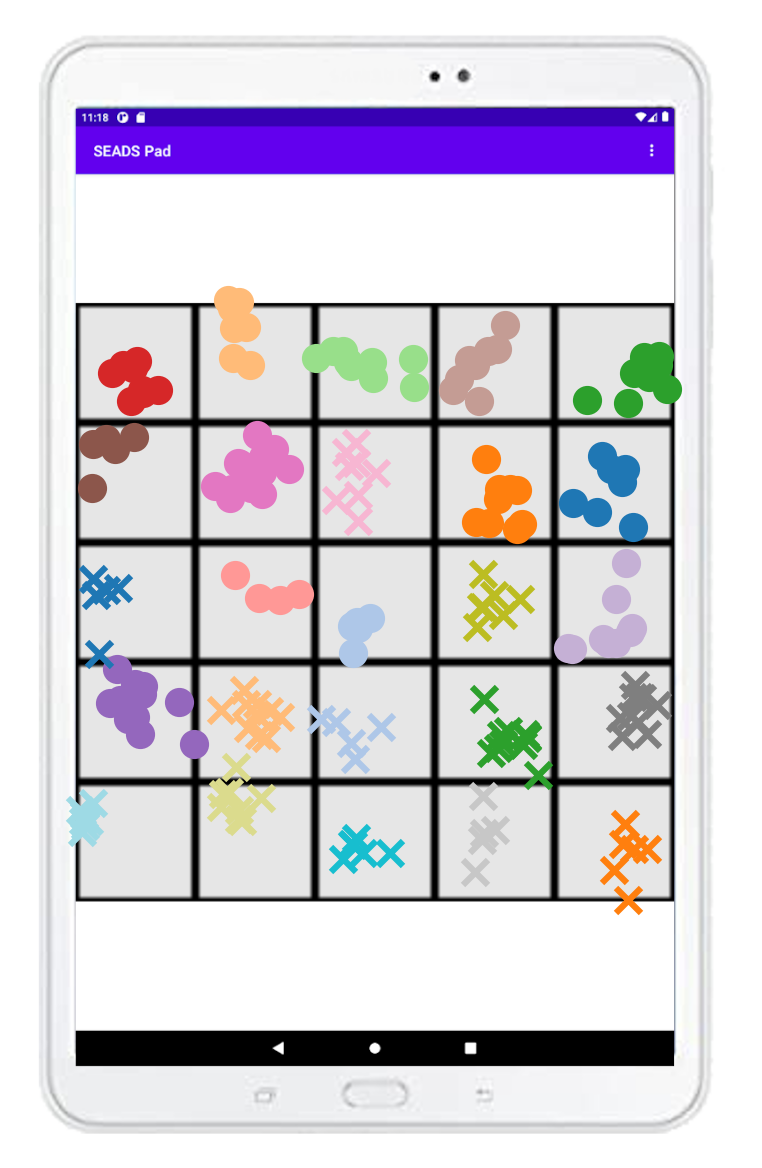}
\caption{Contact points of the last 200 skill executions with randomly sampled skills $k \in \{1,...,25\}$ when training \agent on 5000 environment interactions.}
\label{fig:suppl_robotexp_end}
\end{subfigure}
\hfill
\caption{Real-world setup with a robotic arm (a), on which \agent learns a symbolic forward model on the LightsOut board state and associated low-level skills which relate to pushing locations on the tablet's surface. In (b) and (c) we depict the first 200 and last 200 pushing locations of 5000 pushes used for training in total. While pushing locations are randomly scattered at the beginning of training (b), in the last 200 of 5000 training interactions skills relate to pushing particular fields on the game board (c).}
\label{fig:suppl_robotexp}
\end{figure}

{
In a first variant, the action space of the robotic environment is 2-dimensional, comprising a normalized pushing coordinate $(x, y) \in [0, 1]$ which is translated into a sequence of three commands sent to the robot, setting the position of the end effector to $(x, y, z=0.2)$, $(x, y, z=0)$, $(x, y, z=0.2)$ subsequently. 
To simulate a more realistic gameplay, we do not resample the LightsOut board state or the robots' pose at the beginning of an episode.
We show the setup and behaviour during training in Fig.~\ref{fig:suppl_robotexp}. 
After 5000 environment interactions (corresponding to $\approx 7.5$ hours total training time) we evaluated the SEADS agent’s performance on 100 board configurations (20 per solution depth in $\{1,...,5\}$) and found all of them to be successfully solved by the agent.
}

\subsection{Training on robot with positional displacement actions}
\label{sec:realrobot_posdisplacement}
\begin{figure}[h!]
  \centering
  \includegraphics[width=0.8\linewidth]{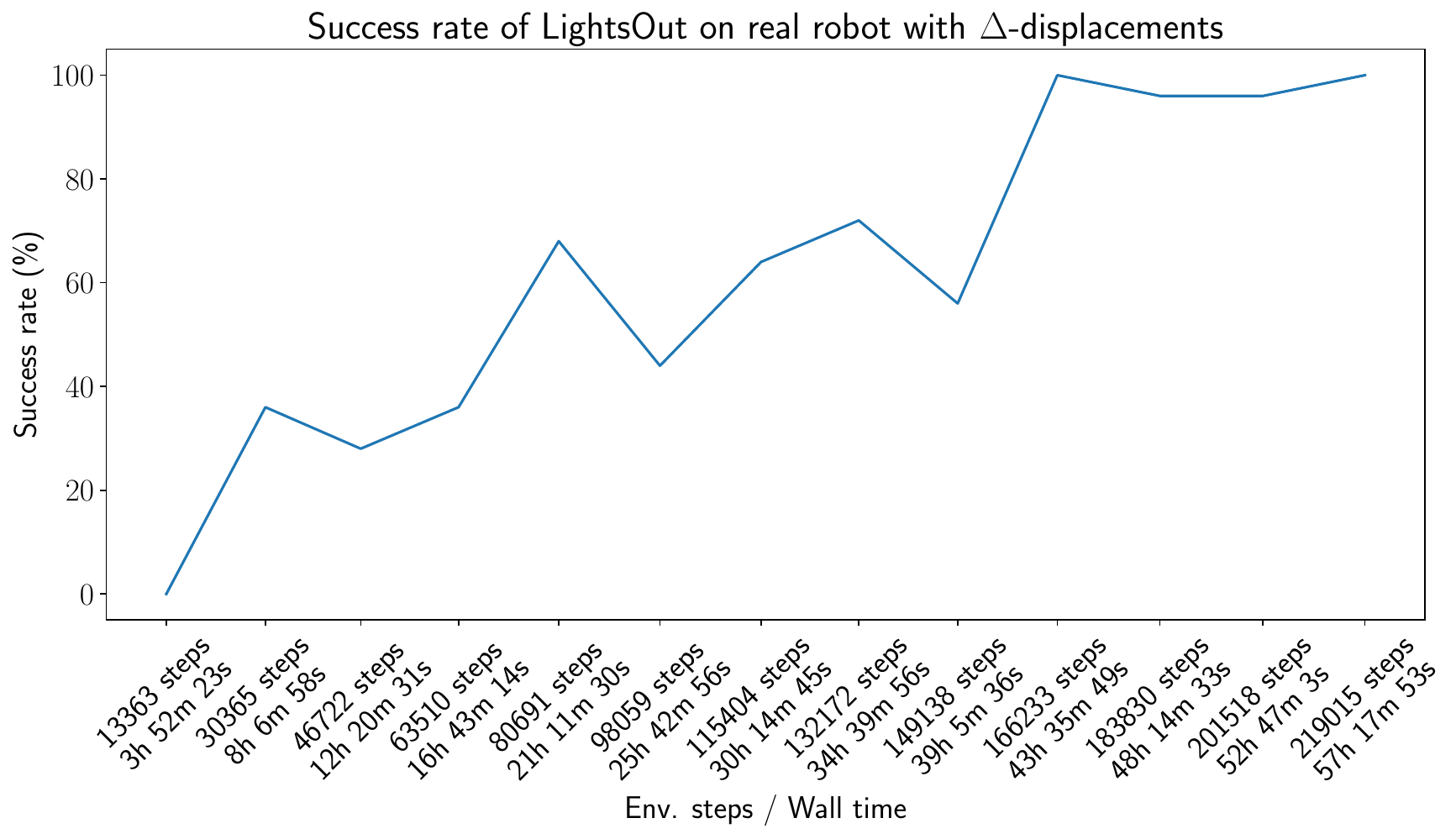} \\
  \caption{Task success rate during the course of training \agent on the \LightsOut game embedded into a physical manipulation scenario with the uArm Swift Pro robotic arm. The arm is controlled by \agent with positional displacement actions (see~\ref{sec:realrobot_posdisplacement}). We evaluate the task performance as success rate on 25 board configurations for each reported step (5 instances per solution depth in $\{1, ..., 5\}$). }
  \label{fig:results_delta_displacements}
\end{figure}

\begin{figure}[h!]
  \centering
  \includegraphics[width=0.8\linewidth]{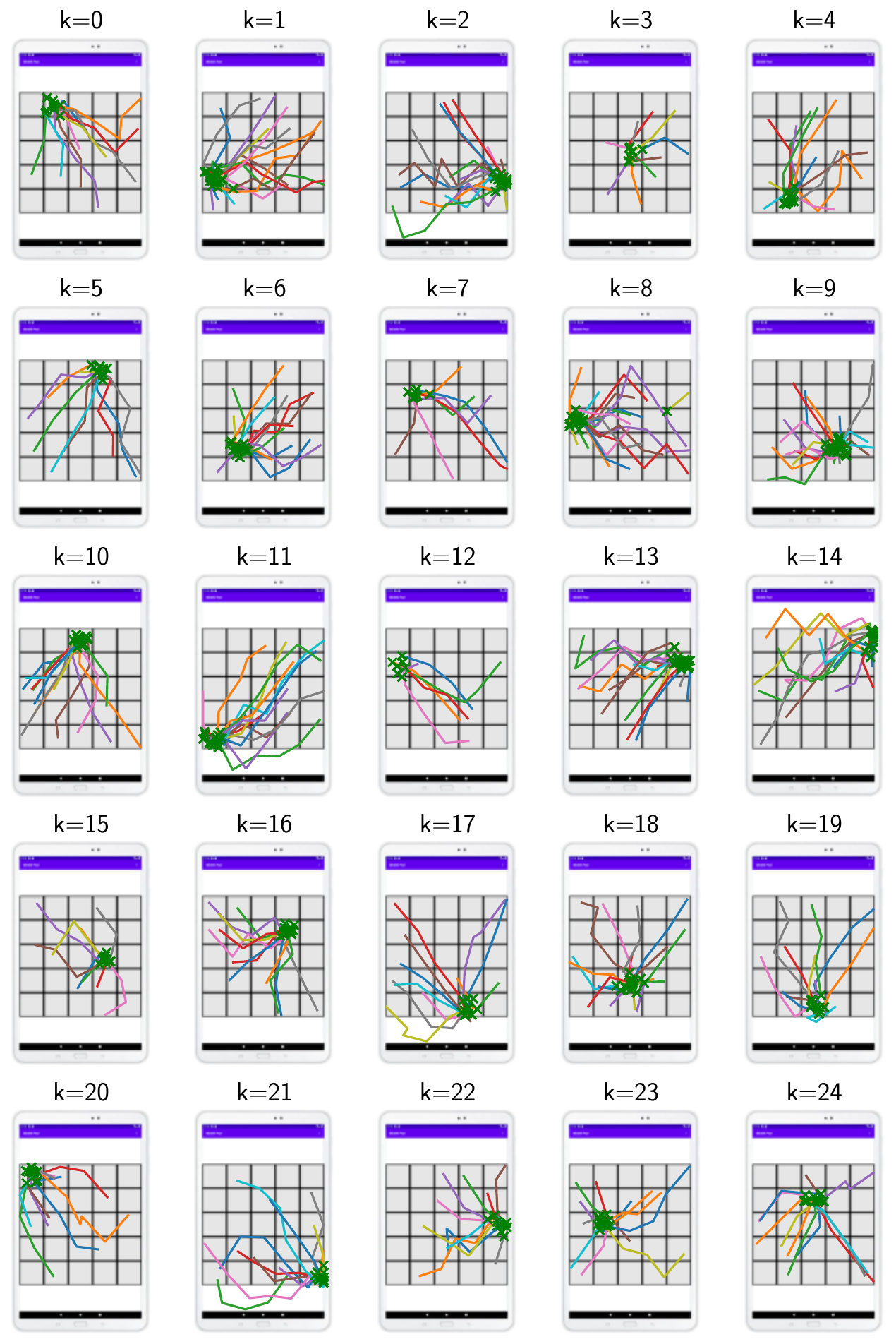} \\
  \caption{Visualization of 320 trajectories executed on the uArm Swift robotic arm with positional displacement actions, after the agent has been trained for $\approx 220k$ environment interactions.  Each subpanel shows trajectories for a specific symbolic action $k$. Green markers indicate push locations. Similar to the other environments, \agent has learned to push individual fields as skills.}
  \label{fig:trajs_delta_displacements}
\end{figure}

{
In this experiment the action space of the environment is 3-dimensional $a = (\Delta x, \Delta y, p)$, with the first two actions being positional displacement actions $\Delta x, \Delta y \in [-0.2, 0.2]$ and the third action $p \in [-1, 1]$ indicating whether a "push" should be executed. The displacement actions  represent incremental changes to the robotic arm's end effector position. In normalized coordinates (see sec.~\ref{sec:suppl_robot_exp}) the end effector is commanded to steer to $(\mathrm{clip}(x+\Delta x, 0, 1), \mathrm{clip}(y+\Delta y, 0, 1), z=0.3)$, where $(x,y)$ are the current coordinates of the end effector. If the push action $p$ exceeds a threshold $p > 0.6$, first the end effector is displaced, followed by a push, which is performed by sending the target coordinates $(x,y,z=0), (x,y,z=0.3)$ to the arm subsequently. In contrast to the first variant in which the \agent agent sends a push location to the agent directly, here the \agent agent has to learn temporally extended skills which first locate the end effector above a particular board field and then execute the push. Therefore, to reach a high success rate on the \LightsOut task, significantly more environment interactions are required compared to the first variant. We observe that a test set of 25 \LightsOut instances (5 per solution depth in $\{1, ..., 5\}$) is solved with a success rate of 100\% after $\approx 165k$ environment interactions, taking in total $\approx 43.5$ hours wall-time to train. We refer to Fig.~\ref{fig:results_delta_displacements} for a visualization of the success rate of \agent over the course of training and to Fig.~\ref{fig:trajs_delta_displacements} for a visualization of skills learned after $\approx 220k$ environment interactions. \emph{We refer to the project page at \url{https://seads.is.tue.mpg.de/} for a video on the real robot experiment.}
}

\section{Extended ablations}
\label{sec:suppl_ablation_analysis}
In this section we present an extended ablation analysis. We refer to sec.~4 of the main paper for a description of the evaluation protocol. 

\subsection{Relabelling}
\begin{figure}[h!]
  \centering
  \includegraphics[width=0.99\linewidth]{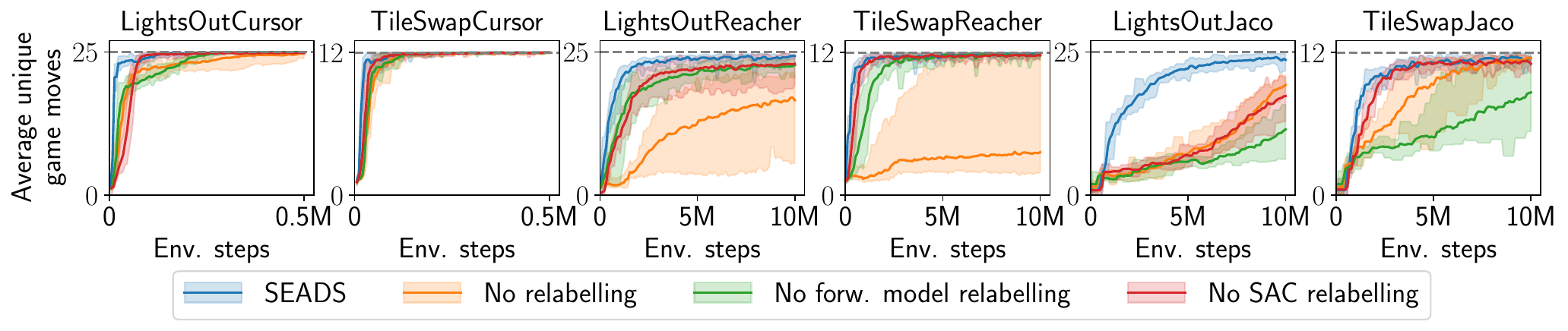}
  \caption{Extended relabelling ablation analysis with 10 trained agents per variant.  The results demonstrate that relabelling both for the forward model and SAC agent training are important for the performance of \agent. See sec.~\ref{sec:suppl_ablation_analysis} for details.}
  \label{fig:suppl_performance_skillquality}
\end{figure}

In this section we provide additional ablations for the episode relabelling. In addition to the "No relabelling" ablation considered in the main paper we investigate not relabelling episodes for the SAC agent only (No SAC relabelling) and not relabelling episodes for training the forward model (No forw. model relabelling). All evaluations are performed on 10 individually trained agents. We refer to Fig.~\ref{fig:suppl_performance_skillquality} for a visualization of the results. The results demonstrate that relabelling both for the forward model and SAC agent training are important for the performance of \agent.

\subsection{Numerical results}
\label{sec:suppl_ablations_numerical}
In Table~\ref{tab:suppl_ablations_table} we present the number of average unique game moves executed as skills by SEADS and its ablations. 
On the simpler Cursor environments, the ablations perform similarly well as SEADS in finding almost all skills. On the Reacher environments, most important for the performance of SEADS is to perform relabelling. On \LightsOutJaco, all ablated design choices are important for the high performance of SEADS.

We also observe that the "More skills" variant (equivalent to \agent, but with $K=30$ for \LightsOut, $K=15$ for \TileSwap) yields a similar number of executed unique game moves as \agent, which is an encouraging result, justifying to over-estimate the number of skills $K$ in situations where it is unknown.  

We perform one-sided Mann-Whitney U tests \cite{mann1947test} to conclude about significance of our results. On each environment, for each of the 10 independently trained SEADS agents, we obtain a set of 10 samples on the average number of detected skills. Analogously, we obtain such a set of 10 samples for every ablation. We aim at finding ablations which either (i) detect significantly more skills than SEADS or (ii) detect significantly less skills than SEADS on a particular environment. We reject null hypotheses for $p < 0.01$.
For (i), we first set up the null hypothesis that the distribution underlying the SEADS samples is stochastically greater or equal to the distribution underlying the ablation samples. For ablations on which this null hypothesis can be rejected it holds that they detect significantly more skills than SEADS. \emph{As we cannot reject the null hypothesis for any ablation, no ablation exists which detects significantly more skills than SEADS.}
For (ii), we set up the null hypothesis that the distribution underlying the SEADS samples is stochastically less or equal to the distribution underlying the ablation samples. \emph{For all ablations there exists at least one environment in which we can reject the null hypothesis to (ii), indicating that all ablations contribute significantly to the performance of SEADS on at least one environment}. The results of the significance test are highlighted in Table~\ref{tab:suppl_ablations_table}.

\begin{table}[h!]
\footnotesize \setlength{\tabcolsep}{0.25em}
\begin{tabular}{lcccccc}
\toprule
 & \multicolumn{2}{c}{Cursor} & \multicolumn{2}{c}{Reacher} & \multicolumn{2}{c}{Jaco}\\
 \cmidrule(lr){2-3} \cmidrule(lr){4-5} \cmidrule(lr){6-7}
 & LightsOut & TileSwap &  LightsOut  & TileSwap & LightsOut  & TileSwap \\
\midrule
SEADS                      &  $24.94 \pm 0.06$ &  $11.99 \pm 0.02$ &   $24.3 \pm 0.28$ &  $11.81 \pm 0.13$ &  $23.64 \pm 1.04$ &  $11.54 \pm 0.27$ \\
No sec.-best norm.         &  $24.74 \pm 0.42$ &  $11.98 \pm 0.03$ &  $24.42 \pm 0.32$ &  $11.78 \pm 0.11$ &  \color{red}$22.28 \pm 0.92$ &   \color{red}$9.26 \pm 1.17$ \\
No nov. bonus              &  $24.83 \pm 0.32$ &  $11.99 \pm 0.03$ &  $23.75 \pm 0.71$ &  $11.81 \pm 0.09$ &  \color{red}$20.24 \pm 1.46$ &  $11.58 \pm 0.29$ \\
No relab.             &  $24.72 \pm 0.39$ &   $12.0 \pm 0.02$ &  \color{red} $16.58 \pm 4.62$ &    \color{red} $3.62 \pm 3.2$ &   \color{red}$19.26 \pm 1.0$ &  $11.53 \pm 0.12$ \\
No forw. mod. relab. &   $24.9 \pm 0.07$ &  $11.99 \pm 0.02$ &  \color{red} $22.73 \pm 0.94$ &  $11.77 \pm 0.28$ &   \color{red}$11.5 \pm 3.47$ &   \color{red}$8.64 \pm 1.83$ \\
No SAC relabelling         &  \color{red} $24.88 \pm 0.09$ &  $11.98 \pm 0.03$ &  \color{red} $22.94 \pm 1.51$ &  $11.76 \pm 0.15$ &  \color{red}$17.32 \pm 2.54$ &  $11.05 \pm 0.48$ \\
\midrule
More skills                &  $24.97 \pm 0.03$ &  $11.99 \pm 0.02$ &  $24.36 \pm 0.36$ &   $11.8 \pm 0.12$ &   $23.9 \pm 0.47$ &  $11.47 \pm 0.41$ \\
\bottomrule
\end{tabular}
\vspace{1em}
\caption{Number of average unique game moves executed as skills by \agent and its ablations, with a maximum of $5\times10^5$ (\Cursor) / $1\times10^7$ (\Reacher, \Jaco) environment interactions (see Sec.~\ref{sec:seads:appendix_checkpointing} for details). The ablated design choices contribute to the performance of \agent especially on the more difficult \Reacher and \Jaco environments. Overestimating the number of skills does not decrease performance ("More skills"). We report mean and standard deviation on 10 independently trained agents. Ablations which perform significantly (one-sided Mann-Whitey U \cite{mann1947test}, $p < 0.01$) worse than \agent are colored {\color{red} red}. We did not find any ablation to perform significantly better than SEADS (including the "More skills" ablation). We refer to sec.~\ref{sec:suppl_ablations_numerical} for details.}
\label{tab:suppl_ablations_table}
\end{table}

\section{Environment details}
\subsection{Cursor environments}
At the beginning of an episode, the position of the cursor is reset to $x \sim \mathrm{Uniform}(0, 1)$, $y \sim \mathrm{Uniform}(0, 1)$. After a game move, the cursor is not reset. Both \LightsOut and \TileSwap board extents are $(x, y) \in [0, 1] \times [0, 1]$.

\subsection{Reacher environments}
At the beginning of an episode, the two joints of the Reacher are set to random angles $\theta \sim \mathrm{Uniform}(0, 2\pi)$, $\phi \sim \mathrm{Uniform}(0, 2\pi)$. After a game move, the joints are not reset.  Both \LightsOut and \TileSwap board extents are $(x, y) \in [-0.15, 0.15] \times [-0.15, 0.15]$. The control simulation timestep is $0.02s$, and we use an action repeat of $2$.

\subsection{Jaco environments}
At the beginning of an episode and after a game move the Jaco arm is randomly reset above the board. The tool's (end effector) center point is randomly initialized to $x \sim \mathrm{Uniform}(-0.1, 0.1)$, $y \sim \mathrm{Uniform}(-0.1, 0.1)$, $z \sim \mathrm{Uniform}(0.2, 0.4)$ with random rotation $\theta \sim \mathrm{Uniform}(-\pi, \pi)$. The \LightsOut board extent is $(x, y) \in [-0.25, 0.25] \times [-0.25, 0.25]$, the \TileSwap board extent $(x, y) \in [-0.15, 0.15] \times [-0.15, 0.15]$. We use a control timestep of $0.1s$ in the simulation.

\section{\agent details}
\label{sec:suppl_details_seads}
In the following, we will present algorithmic and architectural details of our approach.

\subsection{Checkpointing}
\label{sec:seads:appendix_checkpointing}
The number of environment steps varies per model checkpoint due to varying
skill lengths. We report results for budgets of environments steps as the results of model checkpoints being below this threshold.

For the \emph{Skill learning evaluation}, i.e., the average number of detected game moves, we report results on the last checkpoints before reaching $5 \cdot 10^5$ (\Cursor), $10^7$ (\Reacher, \Jaco) env. steps. The results are reported for checkpoints taken after $4.95 \cdot 10^5$ (\Cursor), $9.89 \cdot 10^6$ (\Reacher, \Jaco) env. steps on average.

For the \emph{Planning evaluation}, i.e., the success rate, we again limit the number of environment steps to $5 \cdot 10^5$ (\Cursor), $10^7$ (\Reacher, \Jaco).
We determine results at the last checkpoints for which, for every seed, at least one checkpoint exists with the same or higher number of environment steps (i.e., the last common checkpoints). Effectively, these results are obtained at checkpoints with an average number of $4.76\cdot10^5$ (\Cursor), $9.34\cdot10^6$ (\Reacher, \Jaco) env. steps.

\subsection{Task solution with and without replanning}
One main idea of the proposed \agent agent is to use separate phases of symbolic planning (using the symbolic forward model $q_\theta(z_T \:|\: z_0, k)$) and low-level control (using the skill policies $\pi(a \:|\: s, k)$) for solving tasks. In algorithm~\ref{alg:task_solution_no_repl} and algorithm~\ref{alg:task_solution_repl} we present pseudocode of task solution using planning and skill execution, with and without intermittent replanning.

\begin{algorithm}[tb]
   \caption{Task solution (without replanning). $\operatorname{bfs\_plan}$ denotes breadth-first search over a sequence of skills to transition $z_{(0)}$ to $z^*$, leveraging the symbolic forward model $q_\theta$. Nodes are expanded in BFS via the function $\operatorname{successor}_{q_\theta}:\mathcal{Z}\times\mathcal{K}\to\mathcal{Z}$.}
   \label{alg:task_solution_no_repl}
\begin{algorithmic}
   \STATE {\bfseries Input:} environment $E$, skill-conditioned policy $\pi$, symbolic forward model $q_\theta$, initial state $s_{(0)} = s_0$, symbolic goal $z^*$, symbolic mapping function $\Phi$
   \STATE {\bfseries Output:} boolean  $\mathrm{success}$
   \STATE $N, [k_1, ..., k_N] = \operatorname{bfs\_plan}(q_\theta, z_{(0)}=\Phi(s_{(0)}), z^*)$
   \FOR{$n=1$ {\bfseries to} $N$}
   \STATE $s_{(n)} = \mathrm{apply}(E, \pi, s_{(n-1)}, k_i)$
   \ENDFOR
   \STATE $\mathrm{success} = (\Phi(s_{(N)}) == z^*$)
\end{algorithmic}
\end{algorithm}

\begin{algorithm}[h!]
   \caption{Task solution (with replanning). $\operatorname{bfs\_plan}$ denotes breadth-first search over a sequence of skills to transition $z_{(0)}$ to $z^*$, leveraging the symbolic forward model $q_\theta$. Nodes are expanded in BFS via the function $\operatorname{successor}_{q_\theta}:\mathcal{Z}\times\mathcal{K}\to\mathcal{Z}$.}
   \label{alg:task_solution_repl}
\begin{algorithmic}
   \STATE {\bfseries Input:} environment $E$, skill-conditioned policy $\pi$, symbolic forward model $q_\theta$, initial state $s_{(0)} = s_0$, symbolic goal $z^*$, symbolic mapping function $\Phi$
   \STATE $n \leftarrow 0$
   \REPEAT
       \STATE $N, [k_1, ..., k_N] \leftarrow \operatorname{bfs\_plan}(q_\theta, z_{(n)}=\Phi(s_{(n)}), z^*)$
       \FOR{$i \in \{1, \dots, N\}$}
           \STATE $n \leftarrow n+1$
           \STATE $\hat{z}_{(n)} \leftarrow \operatorname{successor}_{q_\theta}(\Phi(s_{(n-1)}), k_i)$ \COMMENT{Compute the expected symbolic state after applying skill $k_i$}
           \STATE $s_{(n)} \leftarrow \mathrm{apply}(E, \pi, s_{(n-1)}, k_i)$
           \STATE $z_{(n)} \leftarrow \Phi(s_{(n)})$
           \IF{$z_{(n)}\neq \hat{z}_{(n)}$}
           \STATE {\bfseries Break} \COMMENT{If the actual symbolic state differs from the predicted state, we replan}
           \ENDIF
        \ENDFOR
   \UNTIL{$z_{(n)} == z^*$}
\end{algorithmic}
\end{algorithm}

\subsection{Training procedure}
The main training loop of our proposed \agent agent consists of intermittent episode collection and re-labelling for training the skill-conditioned policy $\pi$ using soft actor-critic (SAC, \cite{haarnoja2018soft}) and symbolic forward model (see Algorithm~\ref{alg:training_loop}). For episode collection we first sample a skill from a uniform distribution over skills $k \sim  \mathrm{Uniform}\{1, ..., K\}$ and then collect the corresponding episode. In our experiments we collect 32 episodes per epoch (i.e., $N_\mathrm{episodes} = 32$) for all experiments except the real robot experiment with absolute push positions (sec.~\ref{sec:suppl_robot_exp_abspush}), where we collect 4 episodes per epoch. We maintain two replay buffers of episodes for short-term ($\mathrm{Episodes}_\mathrm{recent}$) and long-term storage ($\mathrm{Episodes}_\mathrm{buffer}$), in which we keep the $N_\mathrm{buffer}=2048$ / $N_\mathrm{recent}=256$ most recent episodes. For training the SAC agent and skill model we combine a sample of $256$ episodes from the long-term buffer and all $256$ episodes from the short-term buffer, comprising the episodes $\mathrm{Episodes}$. These episodes are subsequently passed to the relabelling module. On these relabelled episodes the SAC agent and skill model are trained.
Please see the following subsections for details on episode collection, relabelling, skill model training and policy training.

\begin{algorithm}[h!]
   \caption{\agent training loop}
   \label{alg:training_loop}
\begin{algorithmic}
   \STATE {\bfseries Input:}  Environment $E$ \\
 Number of epochs $N_\mathrm{epochs}$ \\
 Number of new episodes per epoch $N_\mathrm{episodes}$ \\
 Episode buffer size $N_\mathrm{buffer}$ \\
   \STATE {\bfseries Result:} Trained skill-conditioned policy $\pi(a \given s, k)$ and forward model $q_\theta(z_T \given z_0, k)$
   \STATE $\mathrm{Episodes}_\mathrm{buffer} = []$, $\mathrm{Episodes}_\mathrm{recent} = []$
   \FOR{$n_\mathrm{epoch}=1$ {\bfseries to} $N_\mathrm{epochs}$}
   \FOR{$n_\mathrm{episode}=1$ {\bfseries to} $N_\mathrm{episodes}$}
   \STATE Sample $k \sim p(k)$
   \STATE $s_0 = E.\mathrm{reset()}$
   \STATE $\operatorname{Ep} = \mathrm{collect\_episode}(E, \pi, s_0, k)$
   \STATE $\mathrm{Episodes}_\mathrm{buffer}.\mathrm{append}(\operatorname{Ep})$, $\mathrm{Episodes}_\mathrm{recent}.\mathrm{append}(\operatorname{Ep})$
   \ENDFOR
   \STATE $\mathrm{Episodes}_\mathrm{buffer} \leftarrow \mathrm{Episodes}_\mathrm{buffer}$[-$N_\mathrm{buffer}$:] \COMMENT{Keep $N_\mathrm{buffer}$ most recent episodes}
   \STATE $\mathrm{Episodes}_\mathrm{recent} \leftarrow \mathrm{Episodes}_\mathrm{recent}$[-$N_\mathrm{recent}$:]
   \STATE $\mathrm{Episodes} = \mathrm{sample}(\mathrm{Episodes}_\mathrm{buffer}, N=256) \cup \mathrm{Episodes}_\mathrm{recent}$
   \STATE $\mathrm{Episodes}_\mathrm{SM} \leftarrow \mathrm{relabel}(\mathrm{Episodes}, p=1.0)$
   \STATE $\mathrm{update\_skill\_model}(\mathrm{Episodes}_\mathrm{SM})$
   \STATE $\mathrm{Episodes} = \mathrm{sample}(\mathrm{Episodes}_\mathrm{buffer}, N=256) \cup \mathrm{Episodes}_\mathrm{recent}$
   \STATE $\mathrm{Episodes}_\mathrm{SAC} \leftarrow \mathrm{relabel}(\mathrm{Episodes}, p=0.5)$
   \STATE $\mathrm{update\_sac}(\mathrm{Episodes}_\mathrm{SAC})$
   \ENDFOR
\end{algorithmic}
\end{algorithm}

\subsubsection{Episode collection ($\mathrm{collect\_episode}$)}
The operator $\operatorname{collect\_episode}$ works similar to the $\mathrm{apply}$ operator defined in the main paper (see sec. 3). It applies the skill policy $\pi(a_t \given s_t, k)$ iteratively until termination. However, the operator returns all intermediate states $s_0, ..., s_T$ and actions $a_0, ..., a_{T-1}$ to be stored in the episode replay buffers.

\subsubsection{Relabelling ($\mathrm{relabel}$)}
For relabelling we sample a Bernoulli variable with success probability of $p$ for each episode in the $\mathrm{Episodes}$ buffer, indicating whether it may be relabeled. For training the forward model we relabel all episodes ($p=1$), while for the SAC agent we only allow half of the episodes to be relabeled ($p=0.5$). The idea is to train the SAC agent also on \emph{negative} examples of skill executions with small rewards. Episodes in which the symbolic observation did not change are excluded from relabelling for the SAC agent, as for those the agent receives a constant negative reward. All episodes which should be relabeled are passed to the relabelling module as described in sec.~3. The union of these relabelled episodes and the episodes which were not relabeled form the updated buffer which is returned by the $\mathrm{relabel}$ operator.

\subsubsection{SAC agent update ($\mathrm{update\_sac}$)}
In each epoch, we fill a transition buffer using all transitions from the episodes in the $\mathrm{Episodes}_\mathrm{SAC}$ buffer. Each transition tuple is of the form $([s^i_t, k^i], a^i_t, [s^i_{t+1}, k^i], r^i_{t+1})$ where $s$ are environment observations, $a$ low-level actions, $k$ a one-hot representation of the skill and $r$ the intrinsic reward. 
$[\cdot]$ denotes the concatenation operation. 
The low-level SAC agent is trained for $16$ steps per epoch on batches comprising $128$ randomly sampled transitions from the transition buffer. We train the actor and critic networks using Adam \cite{kingma2015adam}. For architectural details of the SAC agent, see sec.~\ref{sec:suppl_seadsarch_sac}.

\subsubsection{Skill model update ($\mathrm{update\_skill\_model}$)}
\label{sec:skill_model_update}
From the episode buffer $\mathrm{Episodes}_\mathrm{SM}$ we sample transition tuples $(z^i_0, k_i, z^i_{T^i})$. The skill model is trained to minimize an expected loss
\begin{equation}
    \mathcal{L} = \mathbb{E}_\mathcal{B} \left[ \: \sum_{i \in \mathcal{B}} L(z^i_0, k_i, z^i_{T^i})  \right]
\end{equation} for randomly sampled batches $\mathcal{B}$ of transition tuples. We optimize the skill model parameters $\theta$ using the Adam \cite{kingma2015adam} optimizer for 4 steps per epoch on batches of size $32$. We use a learning rate of $10^{-3}$. The instance-wise loss $L$ to be minimized corresponds to the negative log-likelihood $L=-\log q_\theta(z^i_{T^i} \given z^i_0, k)$ for symbolic forward models or $L=-\log q_\theta(k \given  z^i_0, z^i_{T^i} )$ for the VIC ablation (see sec.~\ref{sec:suppl_baseline_vic}).
 
\subsection{Architectural details}

\subsubsection{SAC agent}
\label{sec:suppl_seadsarch_sac}
 We use an open-source soft actor-critic implementation \cite{tandon2021sacgithub} in our \agent agent. Policy and critic networks are modeled by a multilayer perceptron with two hidden layers with ReLU activations. For hyperparameters, please see the table below.

\begin{center}
\begin{tabular}{ c c c c c }
\toprule
  \{\LightsOut, \TileSwap\}- & \Cursor & \Reacher & \Jaco & Robot (LightsOut) \\ 
\midrule
 Learning rate & $3\cdot10^{-4}$ & $3\cdot10^{-4}$ & $3\cdot10^{-4}$ & $3\cdot10^{-4}$ \\
 Target smoothing coeff. $\tau$ &  $0.005$ & $0.005$ & $0.005$ & $0.005$ \\
 Discount factor $\gamma$ & 0.99 & 0.99 & 0.99 & 0.99 \\ 
 Hidden dim. & 512 & 512 & 512 & 512\\  
 Entropy target $\alpha$ & 0.1 & 0.01 & 0.01 & 0.1   \\
 Automatic entropy tuning & no & no & no & no   \\
 Distribution over actions & Gaussian & Gaussian & Gaussian & Gaussian \\
\bottomrule
\end{tabular}    
\end{center}

\subsubsection{Symbolic forward model}
\label{sec:suppl_symbolic_forward_model}
Our symbolic forward model models the distribution over the terminal symbolic observation $z_T$ given the initial symbolic observation $z_0$ and skill $k$. It factorizes over the symbolic observation as $q_\theta(z_T \given k, z_0) = \prod_{d=1}^D q_\theta([z_T]_i \given k, z_0) = \prod_{d=1}^D \mathrm{Bernoulli}([\alpha_T(z_0, k)]_i)$ where $D$ is the dimensionality of the symbolic observation. We assume $z \in \mathcal{Z}$ to be a binary vector with $\mathcal{Z} = \{0, 1\}^D$.
The Bernoulli probabilities $\alpha_T(z_0, k)$ are predicted by a learnable neural component. We use a neural network $f$ to parameterize the probability of the binary state in $z_0$ to flip $p_\mathrm{flip} = f_\theta(z_0, k)$, which simplifies learning if the \emph{change} in symbolic state only depends on $k$ and is independent of the current state. Let $\alpha_T$ be the probability for the binary state to be \texttt{True}, then $\alpha_T = (1-z_0)\cdot p_\mathrm{flip} + z_0\cdot (1-p_\mathrm{flip})$. The input to the neural network is the concatenation $[z_0, \mathrm{onehot}(k)]$. We use a multilayer perceptron with two hidden layers with ReLU nonlinearities and 256 hidden units.

\section{SAC baseline}
\label{sec:suppl_baseline_sac}
We train the SAC baseline in a task-specific way by giving a reward of $1$ to the agent if the board state has reached its target configuration and $0$ otherwise. At the beginning of each episode we first sample the difficulty of the current episode which corresponds to the number of moves required to solve the game (solution depth $S$). For all environments $S$ is uniformly sampled from $\{1, ..., 5\}$. For all \Cursor environments we impose a step limit $T_\mathrm{lim}=10\cdot S$, for \Reacher and \Jaco $T_\mathrm{lim}=50\cdot S$. This corresponds to the number of steps a single skill can make in \agent multiplied by $S$. We use a replay buffer which holds the most recent 1 million transitions and train the agent with a batchsize of 256. The remaining hyperparameters (see table below) are identical to the SAC component in \agent; except for an increased number of hidden units and an additional hidden layer (i.e., three hidden layers) in the actor and critc networks to account for the planning the policy has to perform. In each epoch of training we collect 8 samples from each environment which we store in the replay buffer. We performed a hyperparameter search on the number of agent updates performed in each epoch $N$ and entropy target values $\alpha$. We also experimented with skipping updates, i.e., collecting 16 (for $N=0.5$) or 32 (for $N=0.25$) environment samples before performing a single update. We found that performing too many updates leads to unstable training (e.g., $N=4$ for \LightsOutCursor). For all results and optimal settings per environment, we refer to sec.~\ref{sec:suppl_hyperparameter_sac}. For the SAC baseline we use the same SAC implementation from \cite{tandon2021sacgithub} which we use for \agent.

\begin{center}
\begin{tabular}{ c c c c }
\toprule
 \{\LightsOut, \TileSwap\}- & \Cursor & \Reacher & \Jaco \\ 
\midrule
 Learning rate & $3\cdot10^{-4}$ & $3\cdot10^{-4}$ & $3\cdot10^{-4}$ \\
 Target smoothing coeff. $\tau$ &  $0.005$ & $0.005$ & $0.005$ \\
 Discount factor $\gamma$ & 0.99 & 0.99 & 0.99 \\ 
 Hidden dim. & 512 & 512 & 512 \\  
 Entropy target $\alpha$ & \multicolumn{3}{c}{\emph{tuned} (see sec.~\ref{sec:suppl_hyperparameter_sac}) }   \\
 Automatic entropy tuning & no & no & no    \\
 Distribution over actions & Gaussian & Gaussian & Gaussian \\
\bottomrule
\end{tabular}    
\end{center}

\section{HAC baseline}
\label{sec:suppl_baseline_hac}
For the HAC baseline we adapt the official code release \cite{levy2020hacgithub}. We modify the architecture to allow for a two-layer hierarchy of policies in which the higher-level policy commands the lower-level policy with \emph{discrete} subgoals (which correspond to the symbolic observations $z$ in our case). This requires the higher-level policy to act on a discrete action space $\mathcal{A}_\mathrm{high} = \mathcal{Z}$. The lower-level policy acts on the continuous actions space $\mathcal{A}_\mathrm{low} = \mathcal{A}$ of the respective manipulator (\Cursor, \Reacher, \Jaco). To this end, we use a discrete-action SAC agent for the higher-level policy and a continuous-action SAC agent for the lower-level policy. For the higher-level discrete SAC agent we parameterize the distribution over actions as a factorized reparametrizable RelaxedBernoulli distribution, which is a special case of the Concrete \cite{maddison2016concrete} / Gumbel-Softmax \cite{jang2016categorical} distribution. We use an open-source SAC implementation \cite{tandon2021sacgithub} for the SAC agent on both levels and extend it by a RelaxedBernoulli distribution over actions for the higher-level policy.

\subsection{Hyperparameter search}
We performed an extensive hyperparameter search on all 6 environments ([\LightsOut, \TileSwap] $\times$ [\Cursor, \Reacher, \Jaco]) for the HAC baseline. We investigated a base set of entropy target values $\alpha_\mathrm{low}, \alpha_\mathrm{high} \in \{0.1, 0.01, 0.001, 0.0001\}$ for both layers separately. On the \Cursor environments we refined these sets in regions of high success rates. We performed a hyperparameter search on the temperature parameter $\tau$ of the RelaxedBernoulli distribution on the \Cursor environments with $\tau \in \{0.01, 0.05, 0.1, 0.5\}$ and found $\tau=0.1$ to yield the best results. For experiments on the \Reacher and \Jaco environments we then fixed the parameter $\tau=0.1$. We report results on parameter sets with highest average success rate on 5 individually trained agents after $5\times10^5$ (\Cursor) / $1 \times 10^7$ (\Reacher, \Jaco) environment interactions. We refer to sec.~\ref{sec:suppl_hyperparameter_hac} for a visualization of all hyperparameter search results.

\clearpage
\textbf{Parameters for high-level policy} \\[0pt]
\begin{center}
\begin{tabular}{ c c }
\toprule
  & \emph{all environments} \\ 
\midrule
 Learning rate & $3\cdot10^{-4}$ \\
 Target smoothing coeff. $\tau$ &  $0.005$ \\
 Discount factor $\gamma$ & 0.99\\ 
 Hidden layers for actor/critic & 2 \\
 Hidden dim. & 512 \\  
 Entropy target $\alpha_\mathrm{high}$ & \emph{tuned} (see sec.~\ref{sec:suppl_hyperparameter_hac}) \\
 Automatic entropy tuning & no \\
 Distribution over actions & RelaxedBernoulli \\
 RelaxedBernoulli temperature $\tau$ & \emph{tuned} (see sec.~\ref{sec:suppl_hyperparameter_hac}) \\
\bottomrule
\end{tabular}    
\end{center}

\vspace{2em}
\textbf{Parameters for low-level policy} \\[0pt]
\begin{center}
\begin{tabular}{ c c  }
\toprule
 & \emph{all environments} \\ 
\midrule
 Learning rate & $3\cdot10^{-4}$ \\
 Target smoothing coeff. $\tau$ &  $0.005$ \\
 Discount factor $\gamma$ & 0.99\\ 
 Hidden layers for actor/critic & 2 \\
 Hidden dim. & 512 \\  
 Entropy target $\alpha_\mathrm{high}$ & \emph{tuned} (see sec.~\ref{sec:suppl_hyperparameter_hac}) \\
 Automatic entropy tuning & no \\
 Distribution over actions & Gaussian \\
\bottomrule
\end{tabular}    
\end{center}

\section{VIC baseline}
\label{sec:suppl_baseline_vic}
\begin{figure}
  \centering
  \includegraphics[width=0.7\linewidth]{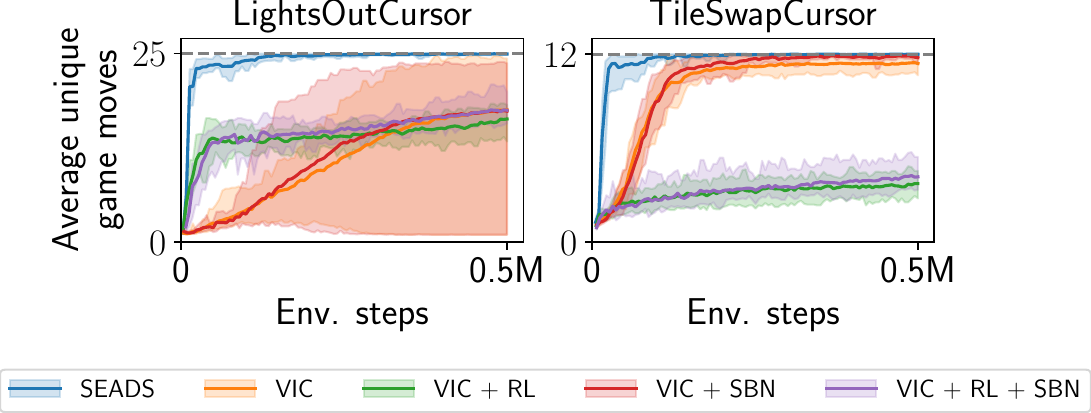} \\
  \caption{Number of discovered skills on the \LightsOutCursor, \TileSwapCursor environments for the \agent agent and variants of VIC \cite{gregor2017_vic}. Only \agent discovers all skills reliably on both environments. See sec.~\ref{sec:suppl_baseline_vic} for details.}
  \label{fig:performance_vic}
\end{figure}

We compare to Variational Intrinsic Control (VIC, \citet{gregor2017_vic}) as a baseline method of unsupervised skill discovery. It is conceptually similar to our method as it aims to find skills such that the mutual information $\mathcal{I}(s_T, k \given s_0)$ between the skill termination state $s_T$ and skill $k$ is maximized given the skill initiation state $s_0$. To this end it jointly learns a skill policy $\pi(s_t \given a_t, k)$ and \emph{skill discriminator} $q_\theta(k \given s_0, s_T)$. We adopt this idea and pose a baseline to our approach in which we model $q_\theta(k \given z_0, z_T)$ \emph{directly} with a neural network, instead of modelling $q_\theta(k \given z_0, z_T)$ indirectly through a forward model $q_\theta(z_T \given z_0, k)$. The rest of the training process including its hyperparameters is identical to \agent. We implement $q_\theta(k \given z_0, z_T)$ by a neural network which outputs the parameters of a categorical distribution and is trained by maximizing the log-likelihood $\log q_\theta(k \given  z^i_0, z^i_{T^i} )$ on transition tuples $(z^i_0, k_i, z^i_{T^i})$ (see sec.~\ref{sec:skill_model_update}). We experimented with different variants of passing ($z^i_0$, $z^i_{T^i}$) to the network: (i) concatenation $[z^i_0$, $z^i_{T^i}]$ and (ii) concatenation with XOR $[z^i_0$, $z^i_{T^i}, z^i_0 \operatorname{XOR} z^i_{T^i}]$. We only found the latter to show success during training. The neural network model contains two hidden layers of size $256$ with ReLU activations (similar to the forward model).
We also evaluate variants of VIC which are extended by our proposed \emph{relabelling scheme} and \emph{second-best reward normalization}. In contrast to VIC, our \agent agent discovers all possible game moves reliably in both \LightsOutCursor and \TileSwapCursor environments, see Fig.~\ref{fig:performance_vic} for details. Our proposed second-best normalization scheme (\emph{+SBN}, sec.~3) slightly improves performance of VIC in terms of convergence speed (\LightsOutCursor) and variance in number of skills detected (\TileSwapCursor). The proposed relabelling scheme (\emph{+RL}, sec.~3) does not improve (\LightsOutCursor) or degrades (\TileSwapCursor) the number of detected skills.

\begin{table}[tbh]
\footnotesize
\begin{center}
\begin{tabular}{ c c c c c c c c c  }
\toprule
    &   \multicolumn{8}{c}{solution depth}\\
    &   1 & 2 & 3 & 4 & 5 & 6 & 7 & 8  \\
\midrule
 LightsOut & 25 & 300 & 2300 & 12650 & 53130 & 176176 & 467104 & 982335  \\ 
 TileSwap & 12 & 88 & 470 & 1978 & 6658 & 18081 & 38936 & 65246  \\ 
\midrule
    &   \multicolumn{8}{c}{solution depth}\\
    &   9 & 10 & 11 & 12 & 13 & 14 & 15 & 16 \\
\midrule
 LightsOut & 1596279& 1935294 & 1684446 & 1004934 & 383670 & 82614 & 7350 & 0 \\ 
 TileSwap & 83000 & 76688 & 48316 & 18975 & 4024 & 382 & 24 & 1 \\ 
\bottomrule
\end{tabular}  
\vspace{1em}
\caption{Number of feasible board configurations for varying solution depths. No feasible board configurations exist with solution depth $>16$.}
\label{tab:suppl_feasible_size}
\end{center}
\end{table}

\section{Environment details}
\label{sec:suppl_env_details}

\subsection{Train-/Test-split}
In order to ensure disjointness of board configurations in train and test split we label each board configuration based on a hash remainder. For the hashing algorithm we first represent the current board configuration as comma-separated string, e.g. $s = \mathrm{"}1,1,0,...,0\mathrm{"}$ for \LightsOut and $s = \mathrm{"}1,0,2,...,8\mathrm{"}$ for \TileSwap. Then, this string is passed through a CRC32 hashing function, yielding the split based on an integer division remainder
\begin{equation}
    \mathrm{split} = \begin{cases}
\mathrm{train} & \operatorname{CRC32}(s) \:\operatorname{mod}\: 3 = 0 \\
\mathrm{test} & \operatorname{CRC32}(s) \:\operatorname{mod}\: 3 \in \{1, 2\} \\
\end{cases}
\end{equation}

\subsection{Board initialization}
We quantify the difficulty of a particular board configuration by its \emph{solution depth}, i.e., the minimal number of game moves required to solve the board. 

We employ a breadth-first search (BFS) beginning from the goal board configuration (all fields \emph{off} in \LightsOut, ordered fields in \TileSwap). Nodes (board configurations) are expanded through applying feasible actions (12 for \TileSwap, 25 for \LightsOut). Once a new board configuration is observed for the first time, its solution depth corresponds to the current BFS step.

By this, we find all feasible board configurations for \LightsOut and \TileSwap and their corresponding solution depths (see table~\ref{tab:suppl_feasible_size}). In table~\ref{tab:suppl_dataset_size} we show the sizes of the training and test split for \LightsOut and \TileSwap environments for solution depths in $\{1,...,5\}$.

\begin{table}
\begin{center}
\begin{tabular}{ c c c c c c c}
\toprule
    &   & \multicolumn{5}{c}{solution depth}\\
    &   & 1 & 2 & 3 & 4 & 5\\
\midrule
 \multirow{3}{*}{LightsOut} & train & 7 & 99 & 785 & 4200 & 17849 \\ 
  & test & 18 & 201 & 1515 & 8450 & 35281  \\ 
  & total & 25 & 300 & 2300 & 12650 & 53130 \\ 
  \midrule
 \multirow{3}{*}{TileSwap} & train & 7 & 31 & 179 & 683 & 2237 \\ 
  & test & 5 & 57 & 291 & 1295 & 4421 \\ 
  & total & 12 & 88 & 470 & 1978 & 6658 \\ 
\bottomrule
\end{tabular}  
\vspace{1em}
\caption{Number of initial board configurations for varying solution depths and dataset splits (train / test).}
\label{tab:suppl_dataset_size}
\end{center}
\end{table}

\section{Results of hyperparameter search on HAC and SAC baselines}
\label{sec:suppl_hyperparameter_search}

\subsection{Results of SAC hyperparameter search}
\begin{figure}[h!]
  \centering
  \includegraphics[width=0.8\linewidth]{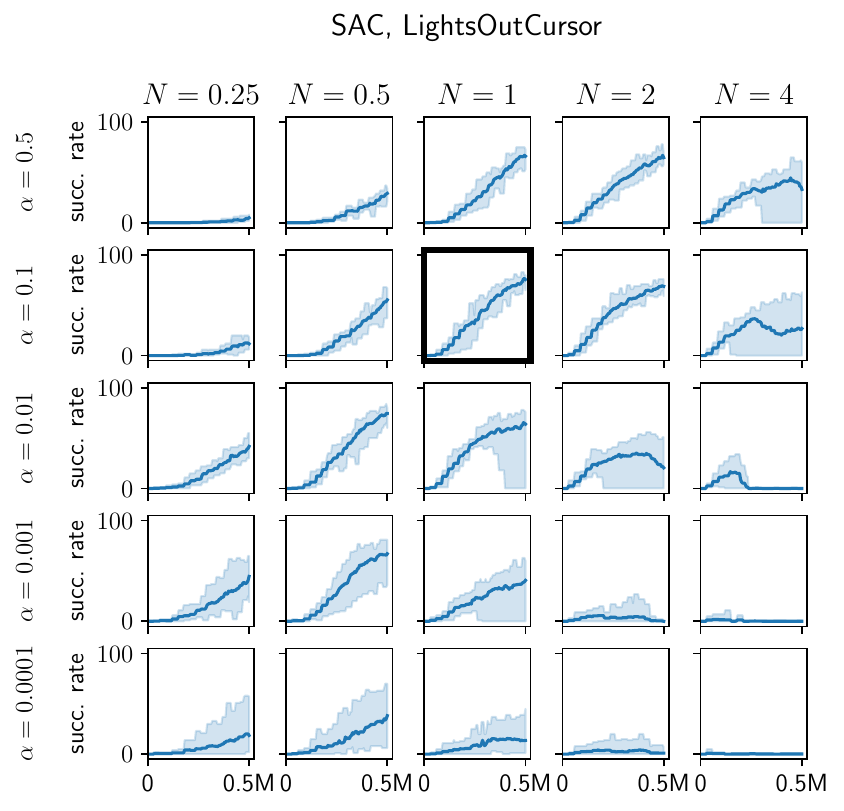}
  \caption{Test performance of SAC agents on the \LightsOutCursor environment for varying number of update steps per epoch ($N$) and parameters $\alpha$. We evaluate 5 individual agents per configuration. The best configuration is marked in \textbf{bold} ($N=1$, $\alpha=0.1$).}
  \label{fig:suppl_sac_hyp_loc}
\end{figure}

\begin{figure}[h!]
  \centering
  \includegraphics[width=0.8\linewidth]{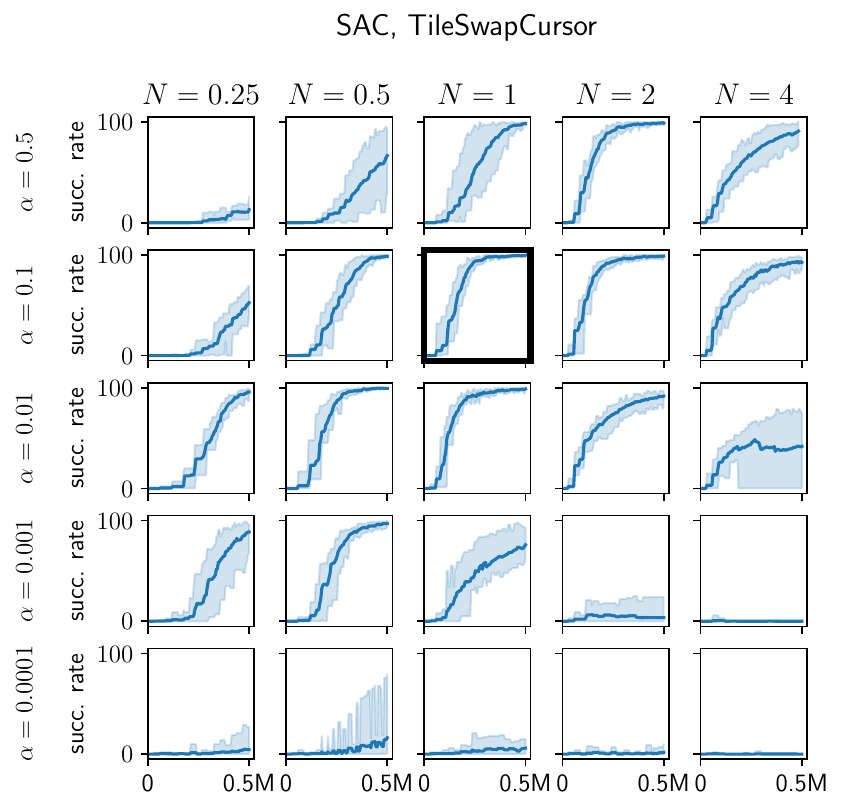}
  \caption{Test performance of SAC agents on the \TileSwapCursor environment for varying number of update steps per epoch ($N$) and parameters $\alpha$. We evaluate 5 individual agents per configuration. The best configuration is marked in \textbf{bold} ($N=1$, $\alpha=0.1$).}
  \label{fig:suppl_sac_hyp_tsc}
\end{figure}

\begin{figure}[h!]
  \centering
  \includegraphics[width=0.55\linewidth]{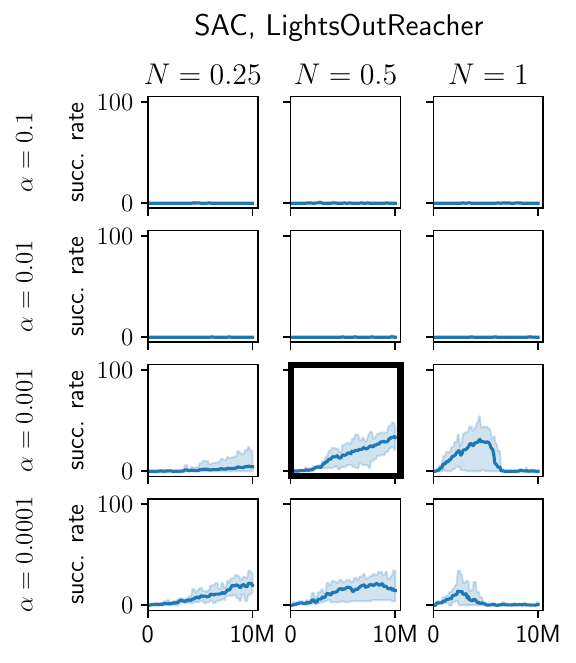}
  \caption{Test performance of SAC agents on the \LightsOutReacher environment for varying number of update steps per epoch ($N$) and parameters $\alpha$. We evaluate 5 individual agents per configuration. The best configuration is marked in \textbf{bold} ($N=0.5$, $\alpha=0.001$).}
  \label{fig:suppl_sac_hyp_lor}
\end{figure}

\begin{figure}[h!]
  \centering
  \includegraphics[width=0.55\linewidth]{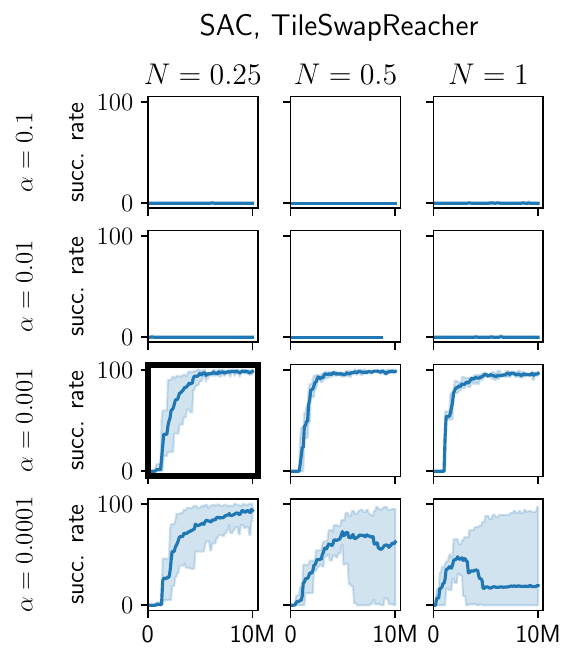}
  \caption{Test performance of SAC agents on the \TileSwapReacher environment for varying number of update steps per epoch ($N$) and parameters $\alpha$. We evaluate 5 individual agents per configuration. The best configuration is marked in \textbf{bold} ($N=0.25$, $\alpha=0.001$).}
  \label{fig:suppl_sac_hyp_tsr}
\end{figure}

\begin{figure}[h!]
  \centering
  \includegraphics[width=0.55\linewidth]{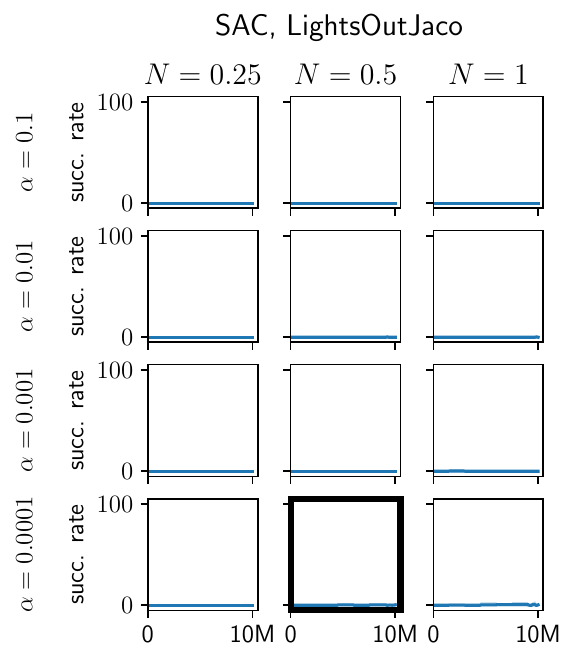}
  \caption{Test performance of SAC agents on the \LightsOutJaco environment for varying number of update steps per epoch ($N$) and parameters $\alpha$. We evaluate 5 individual agents per configuration. The best configuration is marked in \textbf{bold} ($N=0.5$, $\alpha=0.0001$).}
  \label{fig:suppl_sac_hyp_loj}
\end{figure}

\begin{figure}[h!]
  \centering
  \includegraphics[width=0.55\linewidth]{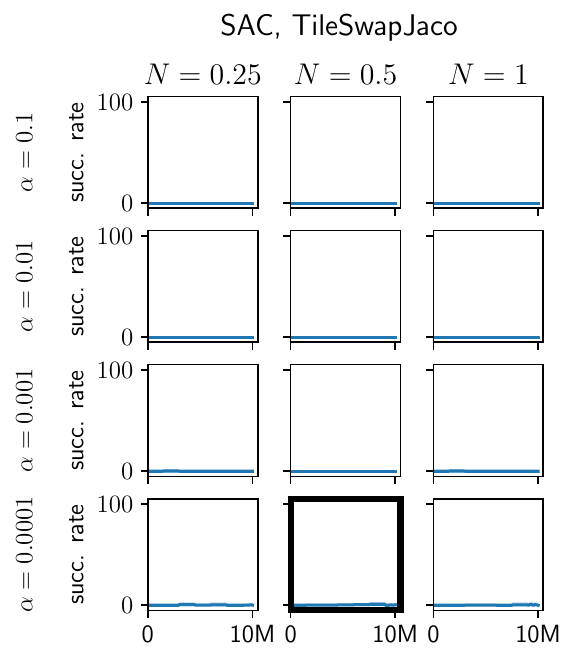}
  \caption{Test performance of SAC agents on the \TileSwapJaco environment for varying number of update steps per epoch ($N$) and parameters $\alpha$. We evaluate 5 individual agents per configuration. The best configuration is marked in \textbf{bold} ($N=0.5$, $\alpha=0.0001$).}
  \label{fig:suppl_sac_hyp_tsj}
\end{figure}

Please see figures~\ref{fig:suppl_sac_hyp_loc}, \ref{fig:suppl_sac_hyp_tsc}, \ref{fig:suppl_sac_hyp_lor}, \ref{fig:suppl_sac_hyp_tsr}, \ref{fig:suppl_sac_hyp_loj}, \ref{fig:suppl_sac_hyp_tsj} for a visualization of the SAC hyperparameter search results.

\label{sec:suppl_hyperparameter_sac}

\clearpage
\subsection{Results of HAC hyperparameter search}
Please see figures~\ref{fig:suppl_hac_hyp_loc}, \ref{fig:suppl_hac_hyp_tsc}, \ref{fig:suppl_hac_hyp_lor}, \ref{fig:suppl_hac_hyp_tsr}, \ref{fig:suppl_hac_hyp_loj}, \ref{fig:suppl_hac_hyp_tsj} for a visualization of the HAC hyperparameter search results.

\label{sec:suppl_hyperparameter_hac}
\begin{figure}[h!]
  \centering
  \includegraphics[width=0.8\linewidth]{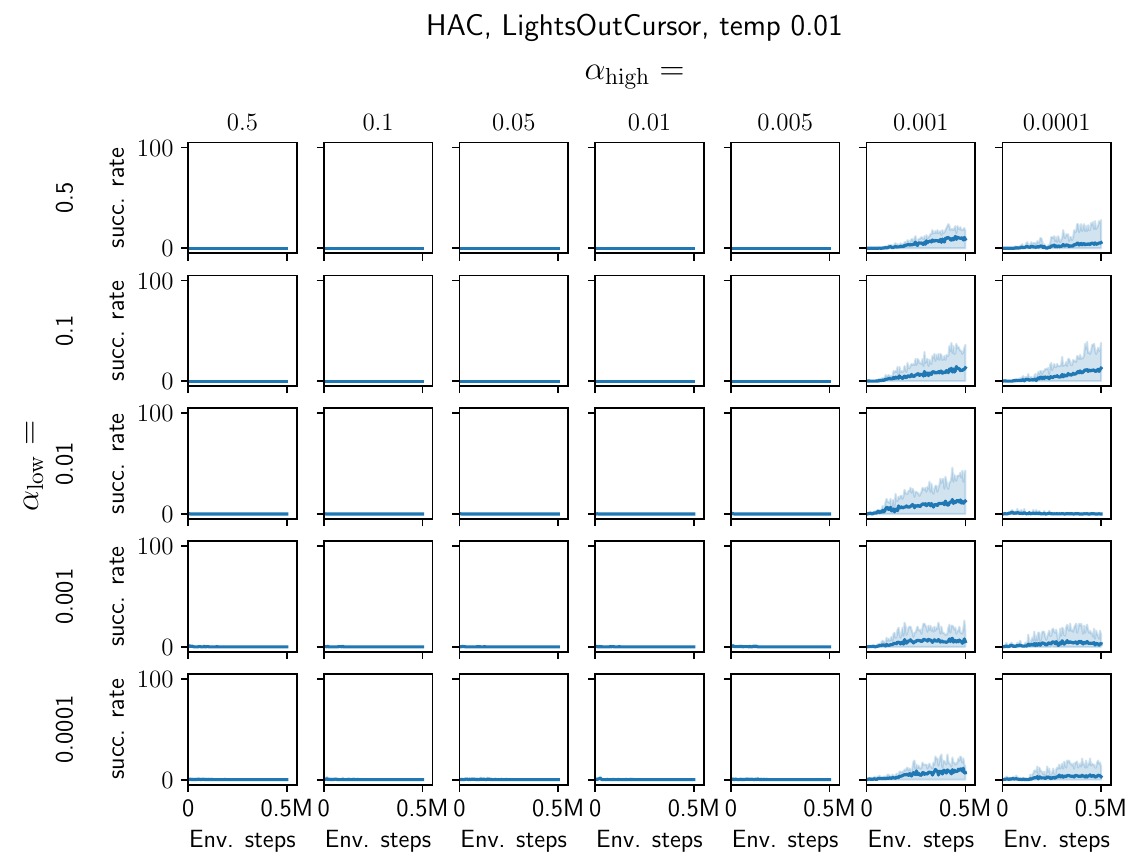} \\[20pt]
  \includegraphics[width=0.8\linewidth]{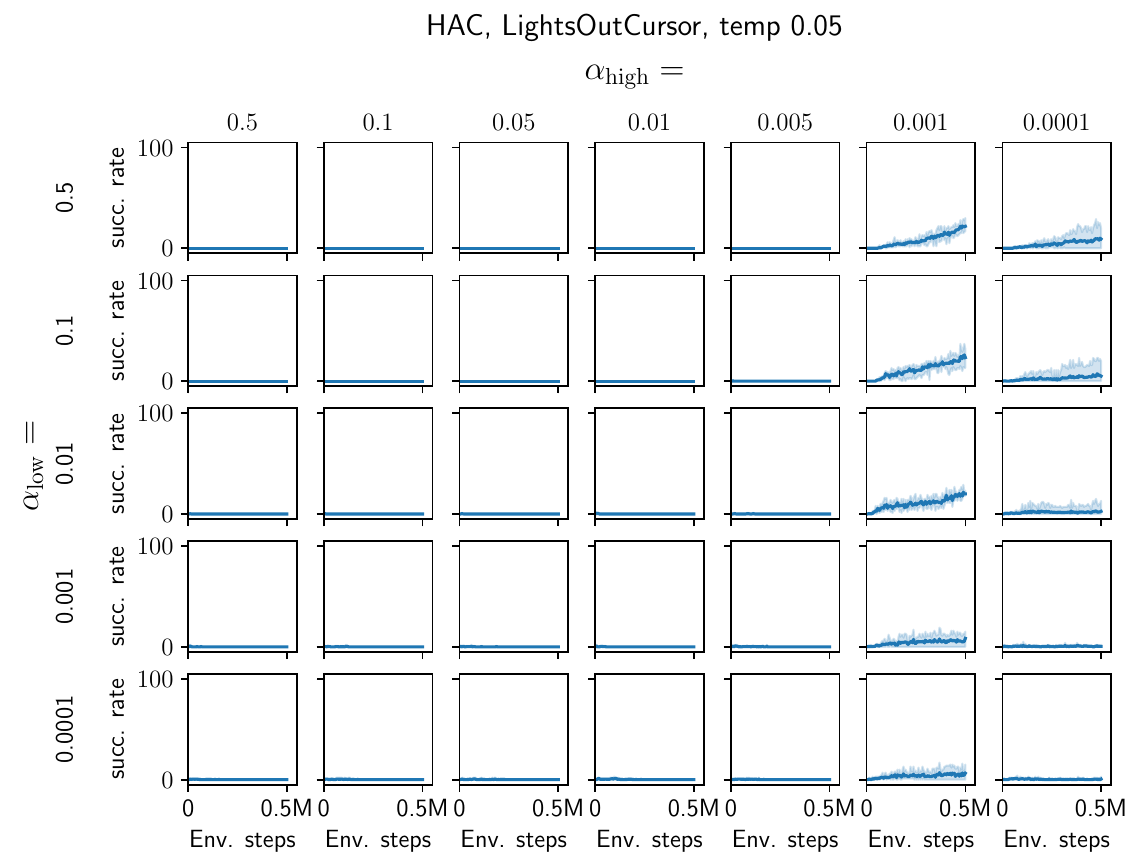} \\[20pt]
\end{figure}

\clearpage

\begin{figure}[h!]
  \ContinuedFloat
  \centering
  \includegraphics[width=0.8\linewidth]{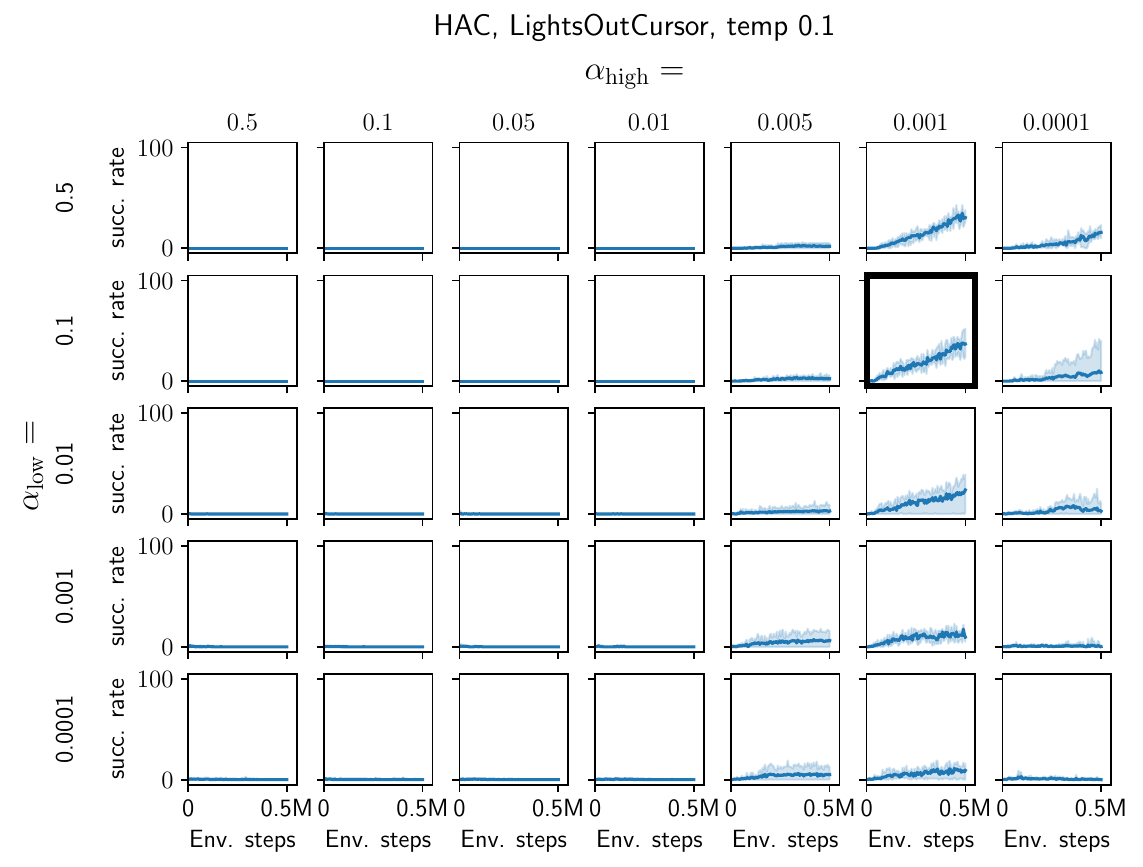} \\[20pt]
  \includegraphics[width=0.8\linewidth]{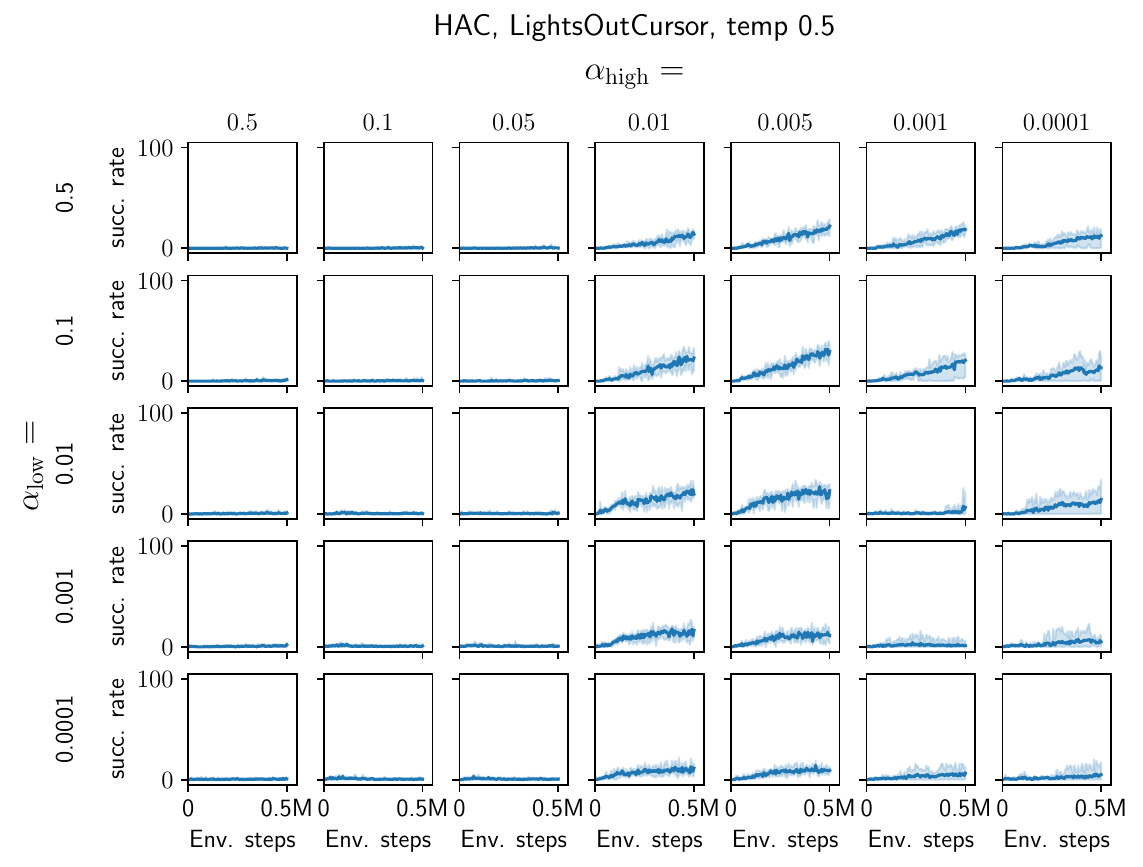} \\[20pt]
  \caption{Test performance of HAC agents on the \LightsOutCursor environment for varying values for RelaxedBernoulli temperature $\tau$ and entropy targets $\alpha_\mathrm{high}, \alpha_\mathrm{low}$. We evaluate 5 individual agents per configuration. The best configuration is marked in \textbf{bold} ($\tau=0.1$, $\alpha_\mathrm{low}=0.1$, $\alpha_\mathrm{high} = 0.001$).}
  \label{fig:suppl_hac_hyp_loc}
\end{figure}

\begin{figure}[h!]
  \centering
  \includegraphics[width=0.8\linewidth]{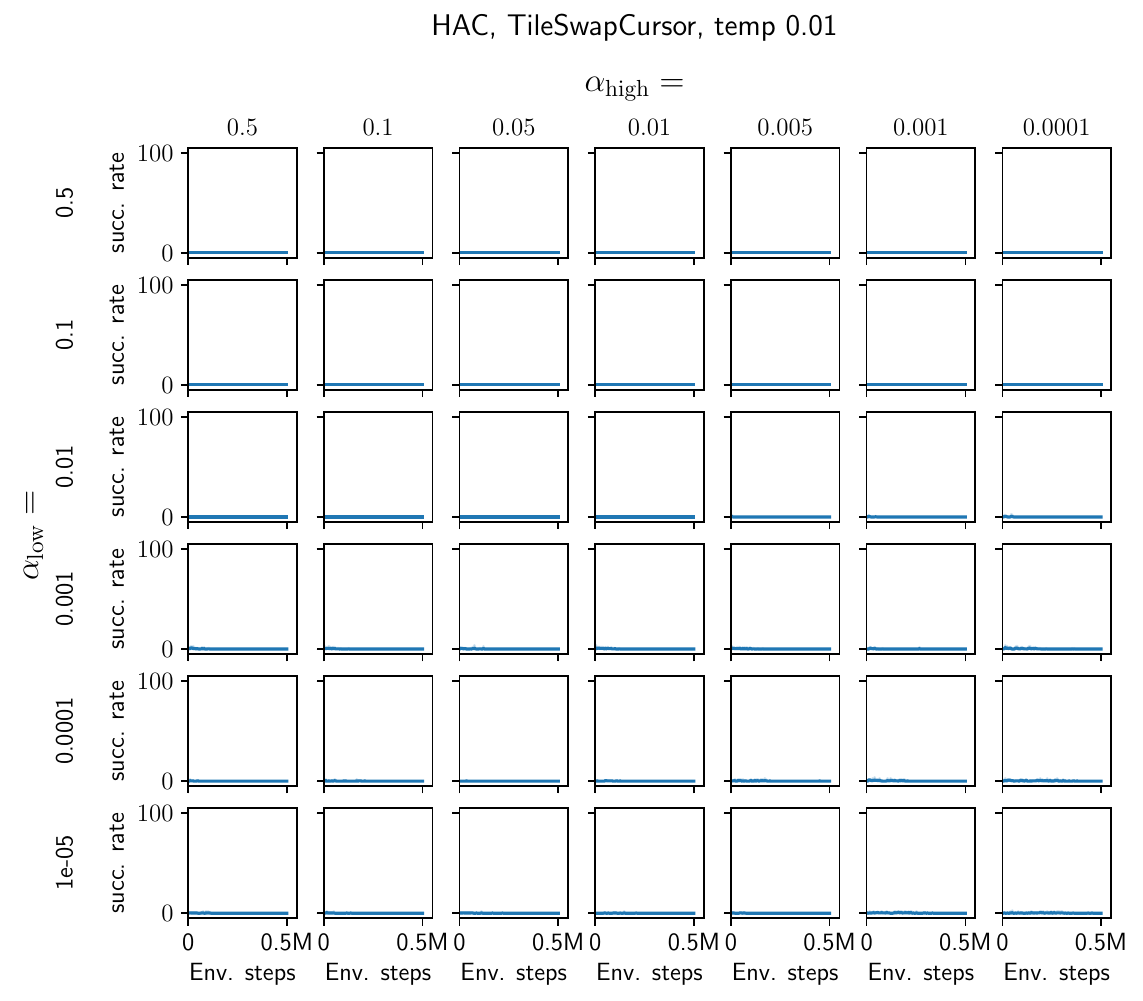} \\[20pt]
  \includegraphics[width=0.8\linewidth]{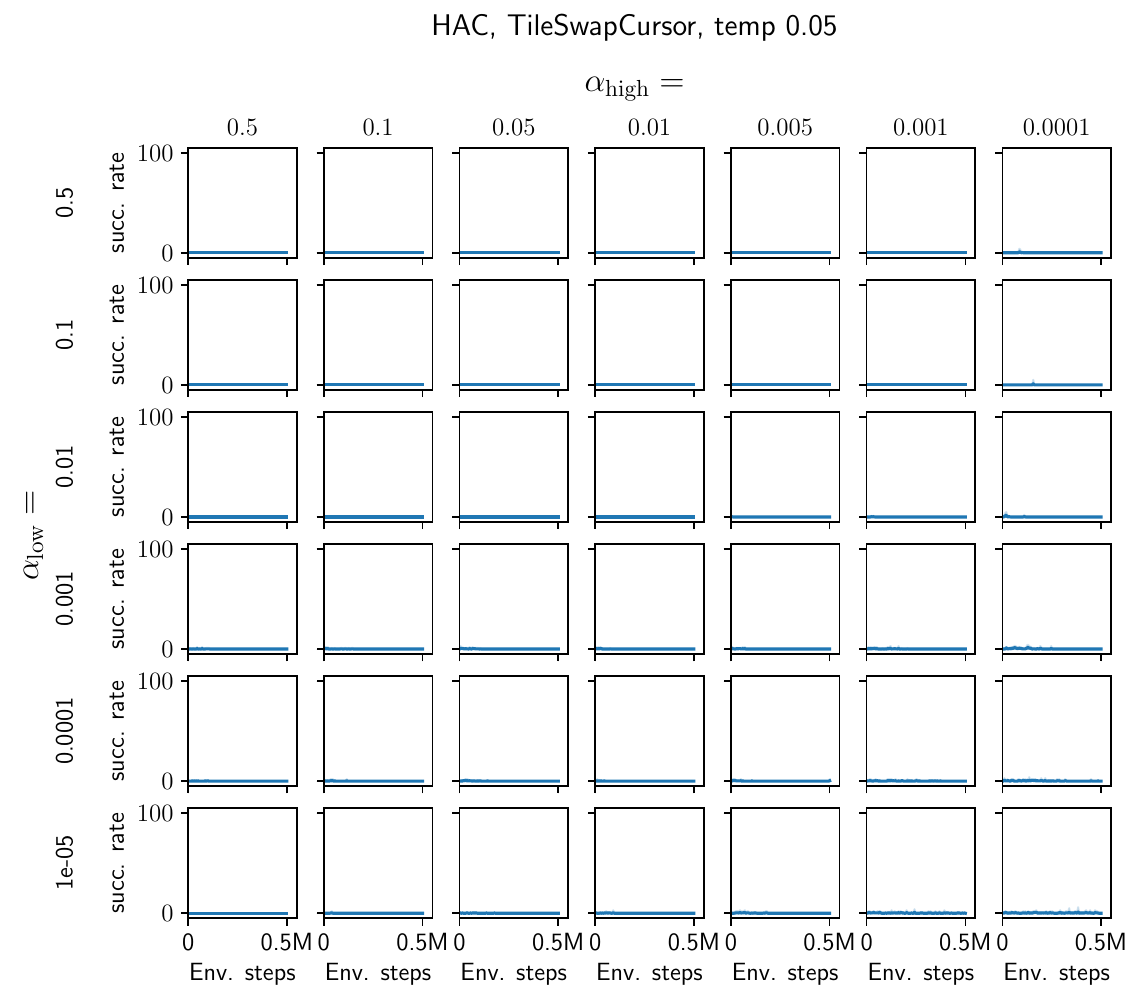} \\[20pt]
\end{figure}

\clearpage

\begin{figure}[h!]
  \ContinuedFloat
  \centering
  \includegraphics[width=0.8\linewidth]{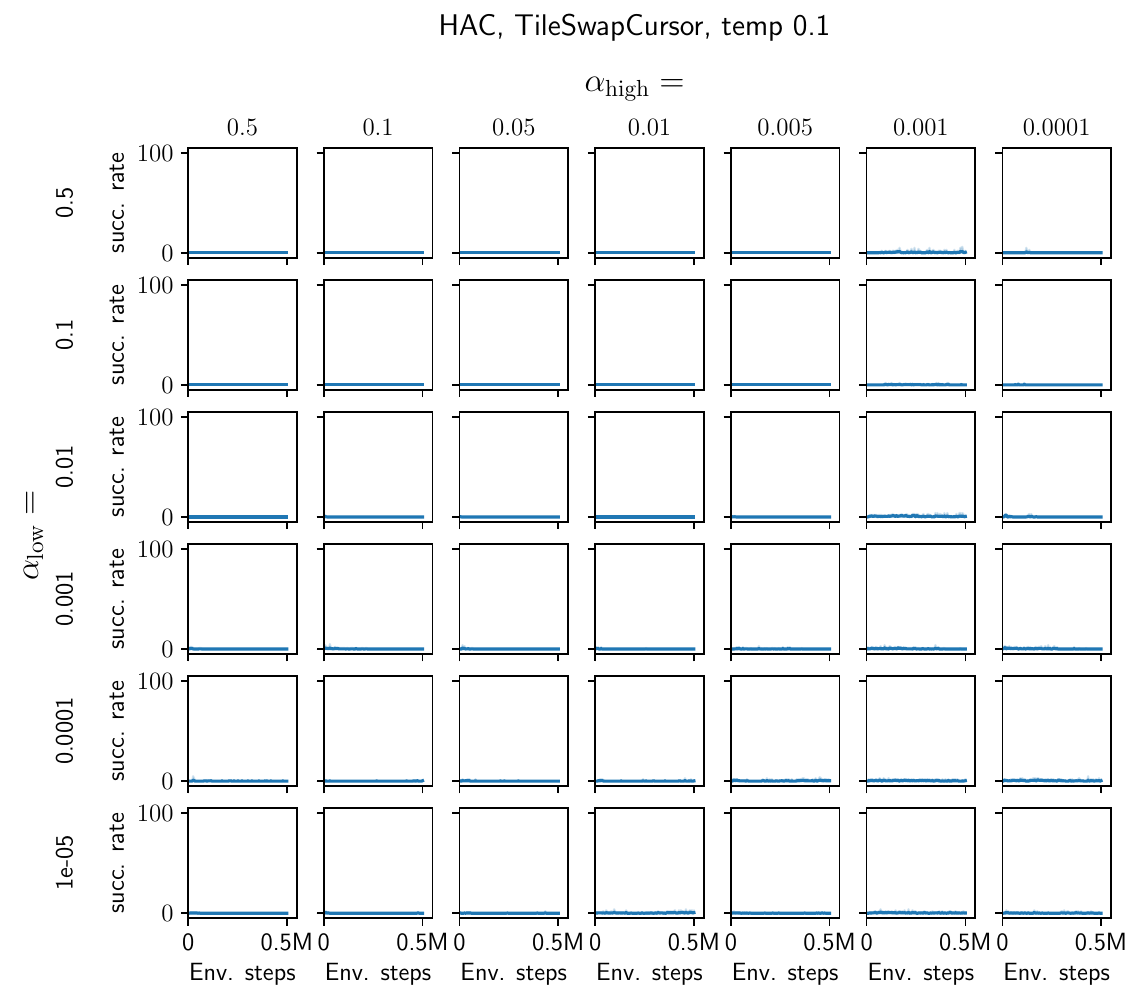} \\[20pt]
  \includegraphics[width=0.8\linewidth]{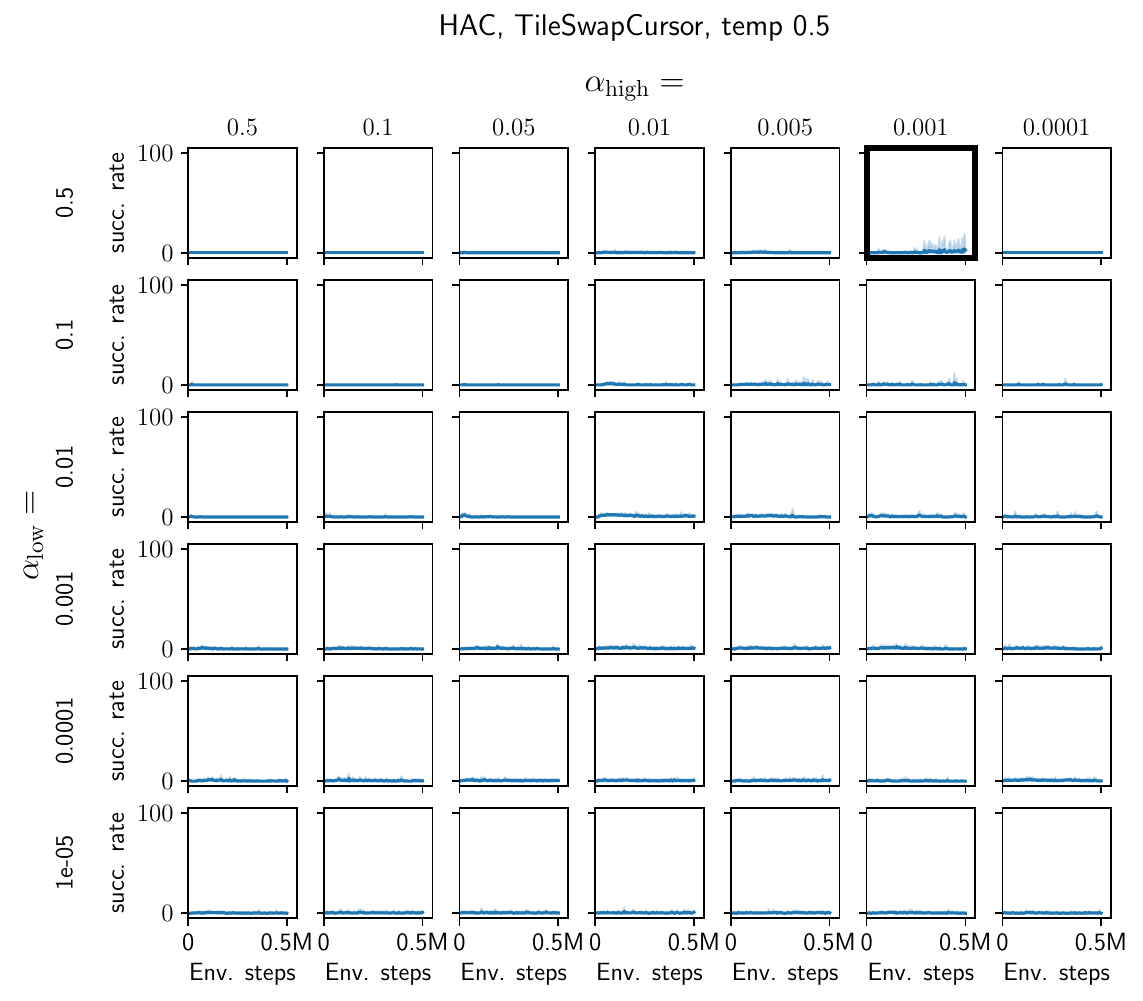} \\[20pt]
  \caption{Test performance of HAC agents on the \TileSwapCursor environment for varying values for RelaxedBernoulli temperature $\tau$ and entropy targets $\alpha_\mathrm{high}, \alpha_\mathrm{low}$. We evaluate 5 individual agents per configuration. The best configuration is marked in \textbf{bold} ($\tau=0.5$, $\alpha_\mathrm{low}=0.5$, $\alpha_\mathrm{high} = 0.001$).}
  \label{fig:suppl_hac_hyp_tsc}
\end{figure}

\clearpage

\begin{figure}[h!]
  \centering
  \includegraphics[width=0.7\linewidth]{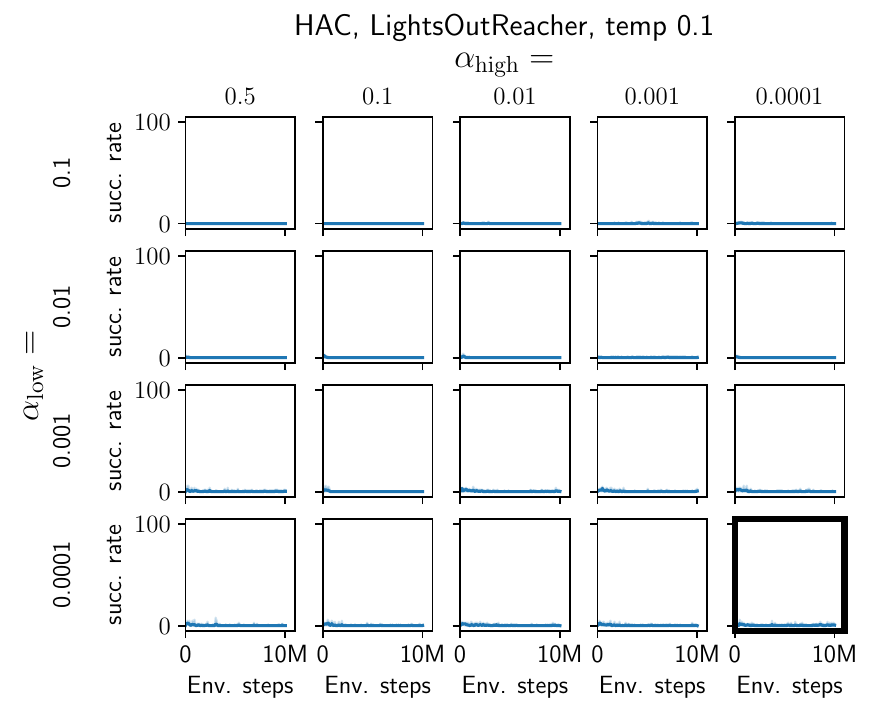}
  \caption{Test performance of HAC agents on the \LightsOutReacher environment for varying values for the entropy targets $\alpha_\mathrm{high}, \alpha_\mathrm{low}$ and fixed RelaxedBernoulli temperature $\tau=0.1$. We evaluate 5 individual agents per configuration. The best configuration is marked in \textbf{bold} ($\tau=0.1$, $\alpha_\mathrm{low}=0.0001$, $\alpha_\mathrm{high} = 0.0001$).}
  \label{fig:suppl_hac_hyp_lor}
\end{figure}

\begin{figure}[h!]
  \centering
  \includegraphics[width=0.7\linewidth]{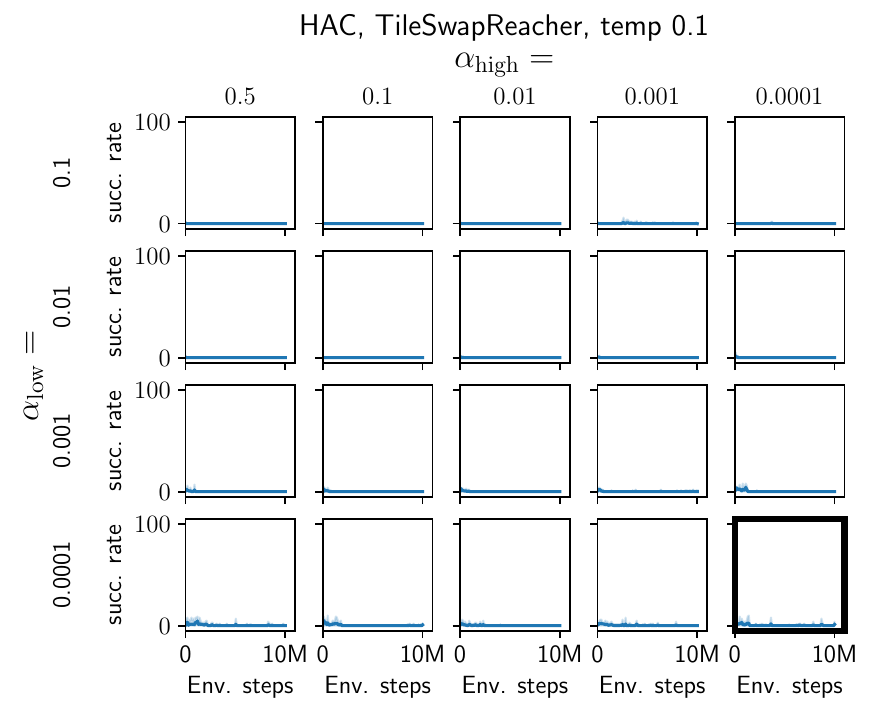}
  \caption{Test performance of HAC agents on the \TileSwapReacher environment for varying values for the entropy targets $\alpha_\mathrm{high}, \alpha_\mathrm{low}$ and fixed RelaxedBernoulli temperature $\tau=0.1$. We evaluate 5 individual agents per configuration. The best configuration is marked in \textbf{bold} ($\tau=0.1$, $\alpha_\mathrm{low}=0.0001$, $\alpha_\mathrm{high} = 0.0001$).}
  \label{fig:suppl_hac_hyp_tsr}
\end{figure}

\clearpage

\begin{figure}[h!]
  \centering
  \includegraphics[width=0.7\linewidth]{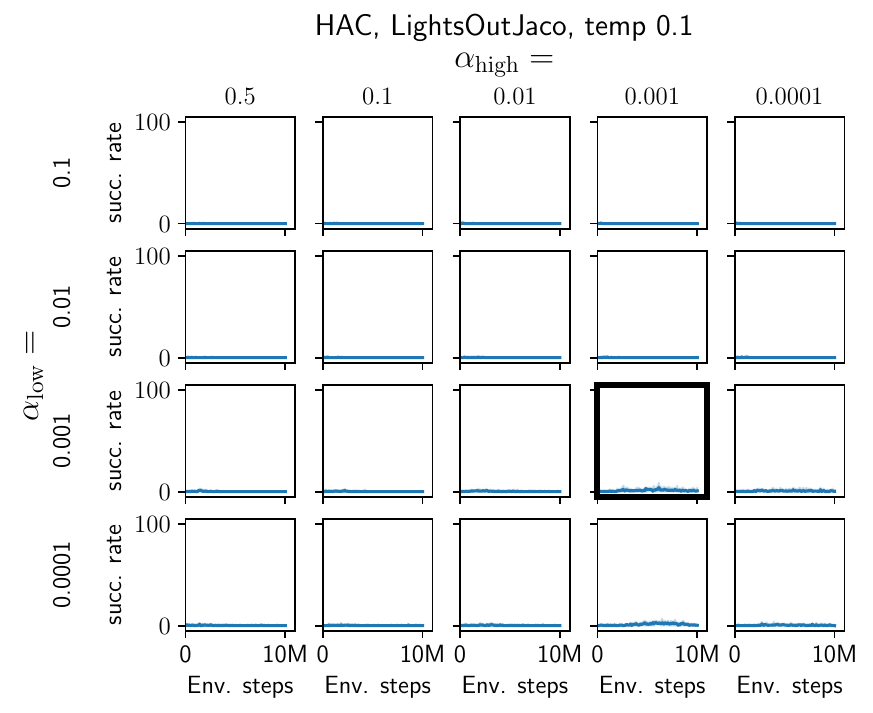}
  \caption{Test performance of HAC agents on the \LightsOutJaco environment for varying values for the entropy targets $\alpha_\mathrm{high}, \alpha_\mathrm{low}$ and fixed RelaxedBernoulli temperature $\tau=0.1$. We evaluate 5 individual agents per configuration. The best configuration is marked in \textbf{bold} ($\tau=0.1$, $\alpha_\mathrm{low}=0.001$, $\alpha_\mathrm{high} = 0.001$).}
  \label{fig:suppl_hac_hyp_loj}
\end{figure}

\begin{figure}[h!]
  \centering
  \includegraphics[width=0.7\linewidth]{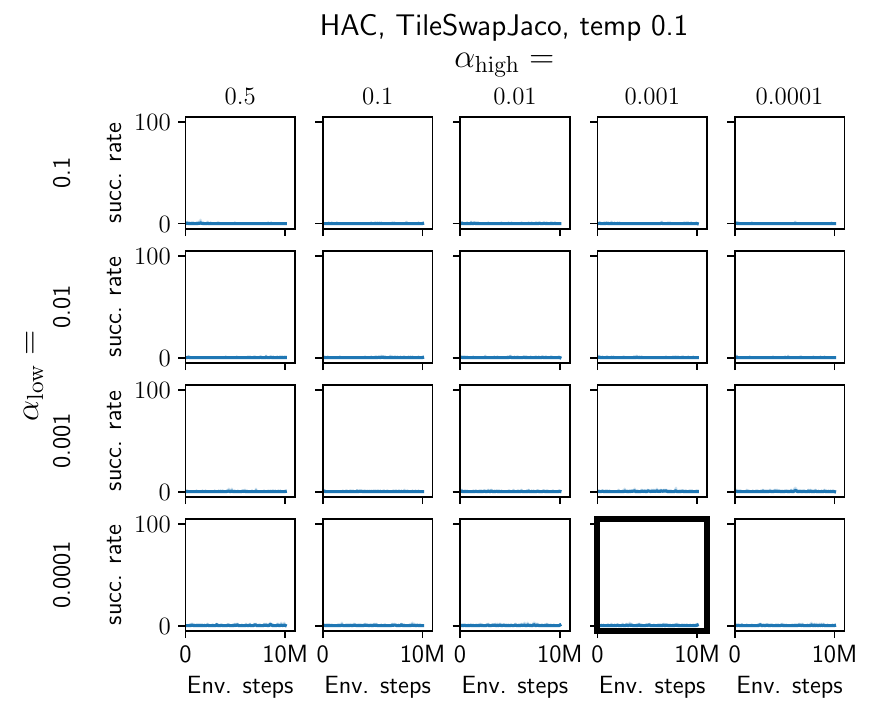}
  \caption{Test performance of HAC agents on the \TileSwapJaco environment for varying values for the entropy targets $\alpha_\mathrm{high}, \alpha_\mathrm{low}$ and fixed RelaxedBernoulli temperature $\tau=0.1$. We evaluate 5 individual agents per configuration. The best configuration is marked in \textbf{bold} ($\tau=0.1$, $\alpha_\mathrm{low}=0.0001$, $\alpha_\mathrm{high} = 0.001$).}
  \label{fig:suppl_hac_hyp_tsj}
\end{figure}

\clearpage

\printbibliography

\end{refsection}

\end{document}